%% file: iclr2026_conference.tex
\documentclass{article} 
\usepackage{iclr2026_conference,times}

\input{math_commands.tex}

\usepackage{hyperref}
\usepackage{url}

\usepackage{amsmath,amssymb,amsthm,amsfonts,mathtools,enumitem}

\newtheorem{theorem}{Theorem}
\newtheorem{lemma}{Lemma}

\newtheorem{proposition}{Proposition}
\newtheorem{remark}{Remark}
\newtheorem{corollary}{Corollary}

\usepackage{bm}

\PassOptionsToPackage{table,dvipsnames,svgnames}{xcolor}
\DeclareMathAlphabet\mathbfcal{OMS}{cmsy}{b}{n}
\newcommand{\ten}[1]{\mathbfcal{#1}}
\newcommand{\mat}[1]{\mathbf{#1}}
\usepackage{pifont}     
\usepackage[table,xcdraw]{xcolor}
\usepackage{booktabs,multirow,makecell,array,colortbl}
\usepackage{tabularx}   
\usepackage{graphicx,wrapfig,float}

\usepackage{enumitem}

\usepackage{microtype,url,hyperref}

\usepackage{nicefrac}
\title{KANO: Kolmogorov–Arnold Neural Operator}

 \author{%
   \makebox[\textwidth][c]{%
     \begin{tabular}{@{}c@{}}
       \multicolumn{1}{c}{\rule{0pt}{1.4em}}\\[-0.2em]  
       \makebox[0.80\textwidth][c]{%
         Jin~Lee$^{1}$\thanks{\texttt{hojin@ucsb.edu}}\hfill
         Ziming~Liu$^{2}$\hfill
         Xinling~Yu$^{1}$\hfill
         Yixuan~Wang$^{3}$} \\[0.8em]
       \makebox[0.75\textwidth][c]{%
         Haewon~Jeong$^{1}$\hfill
         Murphy~Yuezhen~Niu$^{1}$\hfill
         Zheng~Zhang$^{1}$\thanks{\texttt{zhengzhang@ece.ucsb.edu}}} \\[1em]
       \normalfont
         $^{1}$University of California, Santa Barbara\\
         \normalfont
         $^{2}$Massachusetts Institute of Technology\\
         \normalfont
         $^{3}$California Institute of Technology
     \end{tabular}}%
 }

\iclrfinalcopy 
\begin{document}

\maketitle

\begin{abstract}
We introduce \emph{Kolmogorov–Arnold Neural Operator} (KANO), a dual‑domain neural operator jointly parameterized by both spectral and spatial bases with intrinsic symbolic interpretability. We theoretically demonstrate that KANO overcomes the pure-spectral bottleneck of Fourier Neural Operator (FNO): KANO remains expressive over a generic position-dependent dynamics (variable coefficient PDEs) for any physical input, whereas FNO stays practical only to spectrally sparse operators and strictly imposes fast-decaying input Fourier tail. We verify our claims empirically on position-dependent differential operators, for which KANO robustly generalizes but FNO fails to. In the quantum Hamiltonian learning benchmark, KANO reconstructs ground‑truth Hamiltonians in closed-form symbolic representations accurate to the fourth decimal place in coefficients and attains $\approx6\times10^{-6}$ state infidelity from projective measurement data, substantially outperforming that of the FNO trained with ideal full wave function data, $\approx1.5\times10^{-2}$,  by orders of magnitude.
\end{abstract}

\section{Introduction}\label{sec:intro}
\input{intro}

\section{Background}\label{sec:background}
\input{background}

\section{Theoretical Analysis on FNO's Pure-Spectral Bottleneck}\label{sec:fnobottleneck}
\input{fno}

\section{Kolmogorov–Arnold Neural Operator}\label{sec:kano}
\input{kano}

\section{Experimental Results}\label{sec:experiments}
\input{results}

\section{Conclusion}\label{sec:discussion}
\input{discussion}

\section*{Reproducibility Statement}
\input{reproducibility}

\section*{Ethics Statement}
\input{ethics}

\section*{Acknowledgments}
\input{ack}
\bibliographystyle{iclr2026_conference}
\bibliography{iclr2026_conference}

\appendix
\newpage
\section*{Appendix}

\section{Table of Notation}\label{app:notation}
\input{appendix_notation}

\section{Experiment Details}\label{app:impl}
\input{appendix_impl}

\section{Proofs}\label{app:proofs}
\input{appendix_proof}

\section{Additional Details on the Pure-Spectral Bottleneck of FNO}\label{app:high_order}
\input{appendix_high_order}

\section{Computation and Memory Complexity of KANO}\label{app:comp_mem_complexity}
\input{appendix_computation_memory_complexity}

\section{Use of Large Language Model}
\input{appendix_llm}

\end{document}

%% file: math_commands.tex
\usepackage{amsmath,amsfonts,bm}

\def\eqref#1{equation~\ref{#1}}

\def\1{\bm{1}}

\DeclareMathAlphabet{\mathsfit}{\encodingdefault}{\sfdefault}{m}{sl}
\SetMathAlphabet{\mathsfit}{bold}{\encodingdefault}{\sfdefault}{bx}{n}

\DeclareMathOperator*{\argmin}{arg\,min}

%% file: intro.tex
\vspace{-10pt}
Contemporary science and engineering increasingly operate in regimes where the effective dimensionality and complexity of phenomena and data overwhelm human‑designed calibrations and approximations. This motivates data‑centric modeling of governing dynamics from observations~\citep{SciAI1,SciAI2,SciAI3}. For a learned model to be constituted as a scientific law, it should first \emph{generalize} universally over a well-defined domain, and also should be \emph{interpretable} so that the learned representations can be extracted and reused for verification, testing, and downstream simulation. Mathematically, physical dynamics are generalized as operators as they are often formalized through PDEs~\citep{pdephysics1,pdephysics2}. A large and practically important subclass consists of variable coefficient PDEs, in which at least one term has a coefficient that varies by its variables~\citep{vcpde}; we define physical dynamics governed by such PDEs as \emph{position-dependent dynamics}, when one of the variables that varies the coefficient is position. Examples include fluid flow in media with spatially varying viscosity or conductivity~\citep{fluidmechanics}, and the Schr\"odinger equation with a potential that is a function of position operators~\citep{sakurai}. Scientific AI such as operator networks~\citep{no,deeponet} that efficiently approximate a generic position-dependent dynamics with tractable interpretability are therefore valuable, which we recognize the absence and aim to fill the gap in this work.

An operator network approximates an arbitrary mapping between infinite\mbox{-}dimensional function spaces by first encoding functions into finite latent vectors and then learning the latent\mbox{-}to\mbox{-}latent map that represents the target operator~\citep{deeponeterror}. DeepONet of~\citet{deeponet,deeponetoriginal} implements the most general dense operator network where two neural networks learn both encoding and latent mapping directly from data, based on the theoretical foundation laid by~\citet{chenchen}. Fourier Neural Operator (FNO) of~\citet{fno}, on the other hand, hard\mbox{-}codes the encoding as pseudo\mbox{-}spectral projection with its spectrally diagonal kernels. FNO is provably and empirically superior when its hard-coded sparsity is optimal~\citep{fno,uatfno}, but this spectral sparsity becomes maladaptive for position\mbox{-}dependent or otherwise spectrally dense dynamics~\citep{fnoexpressivity,specbfno}. In such cases, the model size required for a target accuracy can grow super-exponentially~\citep{uatfno}, and although the universal approximation guarantee still holds, realistic size FNO may only converge on an in\mbox{-}sample mapping that fails outside the training distribution. Numerous variants of FNO attempted to break this spectral bottleneck. Some have broadened spectral coverage by exploiting factorized~\citep{ffno} or multi\mbox{-}scale~\citep{mscalefno} spectral kernels, and others have injected local spatial kernels alongside the original spectral ones~\citep{ufno,localizedkernelfno,convfno}. Yet all prior works still privilege the spectral basis and cannot achieve optimal sparsity in the spatial basis. 

In parallel, interpretability has recently pivoted around Kolmogorov--Arnold Network (KAN)~\citep{kan,kan2}, whose edges are trainable univariate functions and thus amenable to human inspection. Several works demonstrate data-driven scientific modeling with KAN: \citet{systemid1,systemid2} use KAN for system identification, and \citet{kanode} replace the MLPs in Neural ODEs~\citep{neuralode} with KANs, each reporting symbolic recovery of benchmark equations and parameters. KANs have also been explored within operator networks: \citet{deepokan} employed KANs instead of MLPs in DeepONet and \citet{amfno,efkan} augmented FNO with KANs. Despite performance gains however, prior KAN-based operator networks have not reported symbolic recovery of the learned operator, leaving the avenue of an interpretable operator network largely unexplored.

To address these research gaps, we introduce the \textbf{Kolmogorov–Arnold Neural Operator (KANO)}, an interpretable operator network jointly parameterized in both spatial and spectral bases with KAN sub-networks embedded in a pseudo-differential operator framework~\citep{hormanderPDO,kohnNirenberg}. The key insight is to represent each component of the operator in the basis where it is sparse: differential terms spectrally, localized terms spatially, to achieve the most compact and tractable representation. Our work offers three main contributions to the scientific AI community. 
\begin{itemize} [leftmargin=*]
\vspace{-5pt}
    \item First, we demonstrate the pure-spectral bottleneck of FNO with an illustrative example and theoretically analyze why FNO cannot converge closely as desired to a generic position-dependent dynamics (variable coefficient PDEs) with a practical parameter complexity.
  
    \item Second, we propose a novel framework of KANO that is expressive over a generic position-dependent dynamics with intrinsic symbolic interpretability. We provide theoretical analysis on KANO's dual-domain (spatial and spectral) expressivity along with the empirical evidences of KANO robustly generalizing on unseen input subspace when FNO fails to. 
 
    \item Finally we validate the performance of KANO on some synthetic operators and a quantum simulation benchmark. KANO successfully recovered the closed-form formula accurately to the fourth decimal place in coefficients. Compared to the FNO baseline, KANO used only $0.03\%$ of the model parameters, but achieved an order lower relative loss $\ell_2$ in our synthetic operator benchmarks, and a four-order lower state infidelity in the quantum Hamiltonian learning benchmark.
\end{itemize}
To the best of our knowledge, our work is the first to demonstrate and quantify the symbolic recovery via KAN in operator learning. We shift the paradigm from mere \emph{universal approximation} in operator learning toward the \emph{universal generalization} of an operator network. Different from DeepOKAN~\citep{deepokan} which replaces MLPs with KANs in DeepONet, our work achieves generalization over disjoint out-of-distribution subspace via a novel architecture design.

%% file: background.tex
\vspace{-5pt}
\subsection{Operator Learning and Fourier Neural Operator}\label{sec:fno_background}

\vspace{-5pt}

Operator learning approximates mapping between infinite\mbox{-}dimensional function spaces, $\ten G:\mathcal A\to\mathcal U$,\footnote{$\mathcal A$ and $\mathcal U$ are Banach function spaces (e.g., Sobolev spaces) defined on a bounded domain $D\subset\mathbb{R}^d$.} from function pairs $\{(\mat{a}_i\in\mathcal A,\mat{u}_i=\ten G(\mat{a}_i)\in\mathcal U)\}_{i=1}^N$\footnote{In practice, each function is sampled on a discretized grid in $D$ and stored as a vector.}\citep{no,operator_learning}. An operator network $\ten{G}_\theta$ first encodes input $\mat{a}_i$ via \emph{encoder} $\ten{E}_m:\mathcal A\!\to\!\mathbb C^{m}$ into a latent vector, then learns the \emph{latent map} $\mat{T}_\theta:\mathbb C^{m}\!\to\!\mathbb C^{m'}$ which the output is reconstructed to approximate the label $\mat{u}_i$ via \emph{reconstructor} $\ten{R}_{m'}:\mathbb C^{m'}\!\to\!\mathcal U$: i.e. $\ten G_\theta=\ten R_{m'}\circ \mat{T}_\theta\circ\ten E_m$ \citep{deeponeterror}. For fixed $(\ten E_m,\ten R_{m'})$, 
we can define the projection $\mathbf{\Pi}$ of an operator $\ten{G}$ as
\begin{equation}\label{eq:operator_network}
\mathbf{\Pi}(\ten G)=\ten R_{m'}\circ\widehat{\mat{T}}\circ\ten E_m
\quad\text{where}\quad
\widehat{\mat{T}}\in\argmin_{\mat{T}:\,\mathbb C^{m}\to\mathbb C^{m'}}\big\|\ten G-\ten R_{m'}\circ \mat{T}\circ\ten E_m\big\|.
\end{equation}
DeepONet~\citep{deeponet,deeponetoriginal} learns $\ten{E}_m$, $\ten{R}_{m'}$, and $\mat{T}_\theta$ all with two sub-networks. FNO~\citep{fno}, on the other hand, \emph{hard-codes} $\ten{E}_m$ to be the truncated Fourier transform and $\ten{R}_{m}$ to be its band-limited inverse.

\paragraph{Fourier Neural Operator (FNO).} Let the domain $D\subset\mathbb{R}^d$ be periodic and write the Fourier transform $\ten F$ of function $\mat{a}(\mat{x})$ as $\hat{\mat{a}}(\bm{\xi})$:
\begin{equation}\label{eq:fourier_transform}
    [\ten{F}\mat{a}](\bm{\xi})\;=\;\widehat{\mat{a}}(\bm{\xi})\;=\;\int_{D} \mat{a}(\mat{x})\,e^{-i\,\bm{\omega}\!\cdot\! \mat{x}}\,d\mat{x},\qquad \bm{\omega}=2\pi\bm{\xi}\in\mathbb Z^{d}.
\end{equation}
For fixed set of retained modes $\bm{\xi}_i \in\Xi=\{\bm{\xi}_1,\dots,\bm{\xi}_m\}\subset\mathbb{Z}^d$, truncated Fourier transform $\ten{F}_m:\mathcal A\!\to\!\mathbb C^{m}$ and its band-limited inverse $\ten{F}_m^{-1}:\mathbb C^{m}\!\to\!\mathcal U$ can be defined as: 
\begin{equation}
\ten{F}_m(\mat{a})=[\widehat{\mat{a}}(\bm{\xi}_1),\ldots,\widehat{\mat{a}}(\bm{\xi}_m)],
\quad\quad\ten{F}_m^{-1} (\ten{F}_m\mat{a})(\mat{x})\;=\;\sum_{j=1}^{m}\widehat{\mat{a}}(\bm{\xi}_j)\,e^{2\pi i\,\bm{\xi}_j\cdot \mat{x}},
\end{equation}
with a slight abuse of notation. A single Fourier layer $\ten L_{\text{\tiny{FNO}}}$ of FNO is written as:
\begin{equation}\label{eq:fno_layer}
    \ten{L}_{\text{\tiny{FNO}}}(\mat{a})(\mat{x})\;=\;\sigma\!\left(
\ten{F}_m^{-1}\!\Big(\mat{R}_{\theta_1}(\bm{\xi})\cdot\;\ten{F}_m(\mat{a})(\bm{\xi})\Big)(\mat{x})\;+\;
\mat{W}_{\theta_2} \cdot \mat{a}(\mat{x})
\right)
\end{equation}
with learnable spectral block-diagonal multiplier $\mat{R}(\bm{\xi})$, parametrized linear transformation $\mat{W}$, and point-wise nonlinear activation  $\sigma$. FNO is comprised of iterative $\ten L_{\text{\tiny{FNO}}}$ between lift-up ($\ten P$) and projection ($\ten Q$) networks:
\begin{equation}\label{eq:fno_arch}
   \ten G^{\text{\tiny{FNO}}}_{\theta}(\mat{a})\;=\;\ten{Q}\circ \ten L_{\text{\tiny{FNO}}}^{(\ell)}\circ\cdots\circ \ten L_{\text{\tiny{FNO}}}^{(1)}\circ \ten{P}(\mat{a})\,. 
\end{equation}
In the perspective of the operator network formulation (\ref{eq:operator_network})~\citep{deeponeterror}, FNO hard-codes its encoder $\ten{E}_m$ as $\ten{F}_m$ and reconstructor $\ten{R}_{m}$ as $\ten{F}_m^{-1}$, then learns the latent map $\mat{T}$ by its iterative layers of parametrized linear kernels interleaved by non-linear activations~\citep{uatfno}.

\vspace{-5pt}
\subsection{Kolmogorov--Arnold Network}\label{sec:kan}
\vspace{-5pt}

KAN~\citep{kan,kan2} replaces fixed node activations of traditional MLP with simple sum operations and train the learnable univariate 1D functions $\phi$ on edges. With layer width $n_l\!\to\! n_{l+1}$ and input field $\mat{x}^{(l)}\!\to\! \mat{x}^{(l+1)}$, a KAN layer yields a function matrix $\bm{\Phi}^{(l)}$ at $l^{\text{th}}$ layer as
\begin{equation}\label{eq:kan_arch}
    \mat{x}^{(l+1)}=\bm{\Phi}^{(l)}\!\mat{x}^{(l)}\footnote{Akin to matrix-vector multiplication but follows the third equation Eq.~\ref{eq:kan_arch} instead of row-vector inner product.}, 
\qquad 
\bm{\Phi}^{(l)}=\big[\;\phi^{(l)}_{q,p}(\,\cdot\,)\;\big]_{q=1,\dots,n_{l+1}}^{p=1,\dots,n_l},
\qquad 
x^{(l+1)}_q=\sum^{n_l}_{i=1}\phi^{(l)}_{q,p}(x^{(l)}_p),
\end{equation}
so each output channel is a sum of edgewise transforms of the inputs~\citep{kan,kan2}. In the original KAN each edge function is a spline expansion
\begin{equation}
    \phi^{(\ell)}_{q,p}(t) \;=\; c^{(\ell)}_{q,p,0}\,b(t)\;+\;\sum_{i=1}^{g} c^{(\ell)}_{q,p,i}\,B_i(t),
\end{equation}
with learnable coefficients for a fixed base 1D function $b(\cdot)$ and 1D B--spline basis $\{B_i\}$. Because every $\phi_{q,p}$ is a 1D curve, KANs are directly inspectable and amenable to visualization followed by symbolic regression. On expressivity, \citet{expressivitykan} theoretically prove that KANs match MLPs up to constant depth and width factors; empirically, with appropriate optimization recipes, KANs and MLPs exhibit comparable scaling on PDE and operator benchmarks~\citep{faircomparison}. Hence, swapping a latent MLP for a KAN preserves expressivity while enabling symbolic readout.

%% file: fno.tex
\vspace{-5pt}

This section first illustrates the pure-spectral bottleneck of FNO. Then we provide a theoretical analysis and prove that FNO suffers the curse of dimensionality for position-dependent dynamics.  FNO is proven to have the universal approximation guarantee over any arbitrary non-linear Lipschitz operator~\citep{uatfno,recentfno1}. This section does not disprove the universal approximation ability of FNO; it illustrates the limitation on the \emph{generalization ability} of FNO stemming from its pure-spectral bottleneck on spectrally dense operators.

\subsection{The Pure-Spectral Bottleneck of FNO}\label{sec:fno_example}

\vspace{-5pt}

We consider the 1D quantum harmonic oscillator Hamiltonian as an example:
\begin{equation}
\label{eq:quantum-oscillator}
    \ten{H}a(x) \;=\; -\partial_{xx} a(x) \;+\; x^{2}\cdot a(x).
\end{equation}
Multiplication and differentiation have a dual relationship under the Fourier transform:
\begin{equation}\label{eq:mult_diff_dual}
    \ten{F}[(-\partial_{xx}a)](\xi)\;=\;\xi^{2}\cdot\hat a(\xi),\quad
\qquad 
\ten{F}[(x^{2}\cdot a)](\xi)\;=\;-\,\partial_{\xi\xi}\hat a(\xi).
\end{equation}
In spectral domain, the spatial differential $\partial_{xx}$ is a spectral multiplier $\xi^2$, whereas the spatial multiplier $x^2$ becomes a spectral differential $\partial_{\xi\xi}$. Consider a truncated polynomial basis $\{1,x,x^2,\ldots,x^{n-1}\}$ and a truncated Fourier basis $e_k(\theta)=e^{ik\theta}$, $k=0,\ldots,n-1$, on a periodic domain. In the spatial (polynomial) basis, the map $a(x)\mapsto x^2\cdot a(x)$ acts as a two-step up-shift sparse matrix
\begin{equation}
\mat{S}^{(2)}_n \;:=\;
\begin{bmatrix}
0 & 0 & 1 & 0 & \cdots & 0 \\
0 & 0 & 0 & 1 & \cdots & 0 \\
\vdots & \vdots & \vdots & \ddots & \ddots & 1 \\
0 & 0 & 0 & \cdots & 0 & 0 \\
0 & 0 & 0 & \cdots & 0 & 0
\end{bmatrix},
\label{eq:shift-x2}
\end{equation}
while in the spectral (Fourier) basis it is a dense Toeplitz matrix~\citep{morrison1995spectral}
\begin{equation}
\mat{T}_n[x^2] \;:=\;
\begin{bmatrix}
c_0 & c_{-1} & c_{-2} & \cdots & c_{-n+1} \\
c_{1} & c_0 & c_{-1} & \cdots & c_{-n+2} \\
c_{2} & c_{1} & c_0 & \cdots & c_{-n+3} \\
\vdots & \vdots & \vdots & \ddots & \vdots \\
c_{n-1} & c_{n-2} & c_{n-3} & \cdots & c_0
\end{bmatrix},
\qquad
c_m=\frac{1}{2\pi}\int_{0}^{2\pi}\theta^2 e^{-im\theta}\,d\theta.
\label{eq:toeplitz-x2}
\end{equation}
Thus each term in $\ten{H}$ is sparse in one basis and dense in the other~\citep{morrison1995spectral}.


An FNO layer $\ten{L}_{\text{\tiny FNO}}$ (\ref{eq:fno_layer}) can easily parametrize $-\partial_{xx}$ by taking $\mat{R}(\xi)\approx \xi^2$. However, approximating the dense \emph{off}-diagonals in $\mat{T}_n[x^2]$ to parametrize $x^2$ must rely on the non-linear activation $\sigma(\cdot)$ since $\mat{R}(\xi)$ and $\mat{W}$ are spectrally diagonal and hence incapable of mixing modes. Let $\mat{z}(\mat{u})$ denote the pre-activation for input $\mat{u}$, then the Jacobian of $\ten{L}_{\text{\tiny FNO}}$ at $\mat{u}$ gives the first-order approximation of the learned map and its Fourier transform reveals the spectral off-diagonals of itself as
\begin{equation}\label{eq:fno-jacobian-fourier}
\ten{F}[\mat{J}(\mat{u})](\bm{\xi},\bm{\xi}')
\;=\;
\Big(\ten{F}[\sigma'\!\big(\mat{z}(\mat{u},\cdot)\big)]\,[\,\bm{\xi}{-}\bm{\xi}'\,]\Big)\cdot
\Big(\mat{W} \;+\; \mat{R}_\theta(\bm{\xi}')\Big).
\end{equation}
Therefore, \emph{all} off-diagonals arise from the spectrum of the $\mat{u}$-dependent gate \(\sigma'(\mat{z}(\mat{u},\cdot))\): FNO’s nonlinearity \emph{can} create off-diagonals, but they are tied to the input distribution of \(\mat{u}\). 
This is the pure-spectral bottleneck of FNO: \emph{spectral off-diagonals of a learned FNO are tied to the train subspace, hence FNO can converge only on the in-sample mapping that fails outside the train distribution}\footnote{This issue of out-of-distribution fragility from underspecification is well studied by~\cite{damour}}. See Appendix~\ref{app:high_order} for further detailed discussion expanding to the arbitrary higher order contribution and deep layered FNO.

\subsection{FNO suffers Curse of Dimensionality on Position-Dependent Dynamics}
\label{sec:thm1}
\vspace{-5pt}

As explained previously, \emph{position operator}, $a(x)\mapsto x\cdot a(x)$, is a highly dense Toeplitz map in the spectral basis~\citep{morrison1995spectral}. Based on the Remark 21 \& 22 of~\citet{uatfno}, we prove that any position-dependent dynamics induces super-exponential scaling in FNO size by the desired error bound: \emph{FNO cannot converge closely as desired on a generic position-dependent dynamics with practical model size, hence can only overfit on the in-sample mapping.} We provide Lemma~\ref{lem:position-lower}, that a single position operator already spreads the input spectra too much for FNO to stay practical, and Theorem~\ref{thm:fno-fail}, expanding Lemma~\ref{lem:position-lower} to an arbitrary composition of position operators.

Following from the operator network formulation (\ref{eq:operator_network}), the error estimate of an operator network $\ten{G}_\theta$ approximating the ground-truth operator $\ten{G}$ in an operator norm is bounded as
\begin{equation}\label{eq:operator_network_error}
\|\ten G-\ten G_\theta\|
\;\le\;
\underbrace{\|\ten G-\ten R_{m'}\circ\widehat{\mat{T}}\circ\ten E_m\|}_{\textbf{projection error}: \epsilon_{\text{proj}}}
\;+\;
\underbrace{\|\ten R_{m'}\circ(\widehat{\mat{T}}-\mat{T}_\theta)\circ\ten E_m\|}_{\textbf{latent network error}: \epsilon_{\text{net}}},
\end{equation}
by the triangle inequility. Latent network error $\epsilon_{\text{net}}$ follows the well-established scaling law of conventional neural networks~\citep{hornik1989multilayer,cybenko1989approximation}. Therefore, whether an operator network is efficient in model and sample size to achieve the desired accuracy hinges on the scalability of the projection error $\epsilon_{\text{proj}}$~\citep{deeponeterror,uatfno,datafno}.

\vspace{-5pt}
\paragraph{Reviewing Remark 21 \& 22 of~\citet{uatfno}} $\epsilon_{\text{proj}}$ of FNO is governed by the Fourier tail, the sum of Fourier coefficients outside the retained spectrum $\Xi$ truncated by width $m$~\citep{spectraltxt,trefethen1999spectral}: to achieve the desired $\epsilon_{\text{proj}}$ with practical $m$, both input and output Fourier tails must decay algebraically or faster. However, even in the optimal case of the band-limited input, if the ground\mbox{-}truth operator is spectrally dense to spread out the input spectra and induce algebraic or slower decay in output Fourier tail, $m$ must scale at least polynomially to suppress $\epsilon_{\text{proj}}$: $m \sim  \mathcal{O}\!\bigl(\epsilon_{\text{proj}}^{-1/s}\bigr)$ where $s$ is a geometric constant. Meanwhile, as the latent mapping would be also dense, the size of the latent network, $\mathcal{N}_{\text{net}}$, follows the canonical polynomial neural scaling~\citep{reluscaling,tanhscaling} by the desired $\epsilon_{\text{net}}$ with its width $m^d$ ($d$ is the input domain dimension) as the exponent: $\mathcal{N}_{\text{net}} \sim \mathcal{O}\!\bigl(\epsilon_{\text{net}}^{-m^d}\bigr)$. Consequently, this results as the super-exponential scaling in the latent network size $\mathcal{N}_{\text{net}} \sim \mathcal{O}\!\bigl(\epsilon_{\text{net}}^{-\epsilon_{\text{proj}}^{-d/s}}\bigr)$ even with the optimal band-limited input: \emph{scaling width $m$ to suppress $\epsilon_{\text{proj}}$ explodes $\mathcal{N}_{\text{net}}$ to achieve the desired $\epsilon_{\text{net}}$ for a generic dense operator.} 

\begin{lemma}[\textbf{Position operator elongates Fourier tail}]
\label{lem:position-lower}
A single position operator, spatial multiplier by $x$, induces algebraic decay in output Fourier tail when the input is band-limited.
\vspace{-10pt}
\begin{proof}[Sketch of proof.]
Position operator is kernel  $\widehat{x}(\xi)\!\propto\!1/\xi$ in spectral basis. Hence, every mode outside the input spectrum picks up a coefficient of size $\sim 1/|\xi|$, ending up as $|\hat v(\xi)|\gtrsim 1/|\xi|$ in the output spectrum. See Appendix~\ref{app:lem1proof} for restatement and full proof.
\end{proof}
\end{lemma}
\begin{theorem}[\textbf{Curse of dimensionality on position operators} ]
\label{thm:fno-fail}
Any arbitrary composition of position operators requires FNO to scale super-exponentially on its model size by the desired accuracy.
\vspace{-10pt}
\begin{proof}[Sketch of proof.]
Iteratively apply Lemma~\ref{lem:position-lower}, then any arbitrary composition of position operators induce algebraic or slower decay in output Fourier tail even for the optimal band-limited input. This results in super-exponential scaling of latent network size by the desired error as discussed above~\citep{uatfno}.  See Appendix~\ref{app:thm1proof} for restatement and full proof.
\end{proof}
\end{theorem}

\begin{remark}
    What is missing from the upper bound analysis of FNO by~\citet{uatfno} is the effect of the wide lift-up and projection networks. For the \emph{generalization guarantee} arguments of this work, the upper bound analysis is still sound, yet it should be clarified that the theoretical analysis on the role of the lift-up and projection networks is yet an open research question. Recent studies by~\citet{recentfno1,recentfno2} provide better view on it with lower bound analysis.
\end{remark}

%% file: kano.tex
\vspace{-5pt}
Motivated by the pure-spectral bottleneck of FNO, we propose the Kolmogorov-Arnold Neural Operator (KANO), an operator network capable of converging closely as desired on a generic position-dependent dynamics with practical model size. We first introduce the KANO architecture, and provide theoretical analysis on its dual-domain expressivity in the following section.

\subsection{KANO Architecture}\label{sec:kanoarch}
\vspace{-5pt}

\begin{figure}[t]
  \centering
  \vspace{-20pt}
  \includegraphics[width=\linewidth]{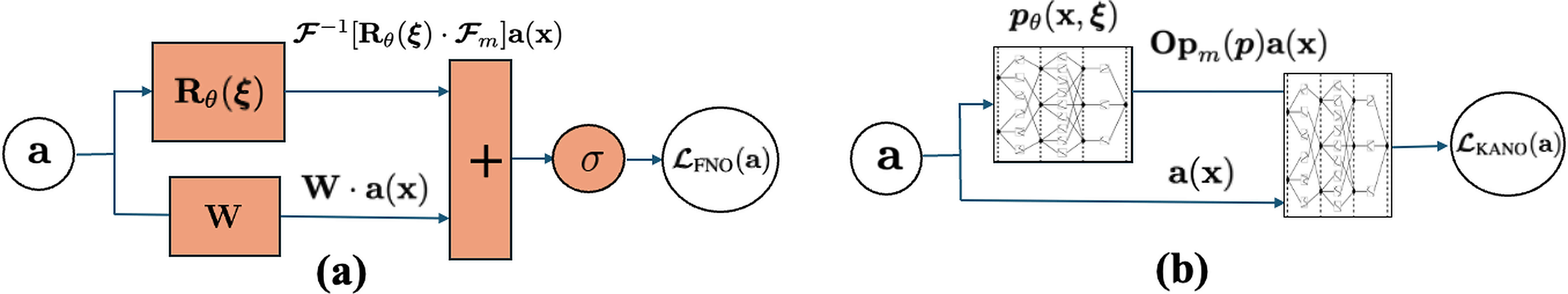}
  \caption{\textbf{(a)} $\ten{L}_{\tiny{\text{FNO}}}$ architecture. \textbf{(b)} $\ten{L}_{\tiny{\text{KANO}}}$ architecture. }
  \label{fig:kano}
  \vspace{-10pt}
\end{figure}

KANO utilizes an iterative structure of KANO layers $\ten{L}_{\text{\tiny{KANO}}}$ to learn the unknown operator, akin to FNO. However, KANO excludes the wide lift-up and projection networks to maximize tractability since it is known that wide KANs are fragile to symbolic recovery~\citep{kanguide}:
\begin{equation}\label{eq:kano-arch}
  \ten G^{\text{\tiny{KANO}}}_{\theta}\;=\;\ten L_{\text{\tiny{KANO}}}^{(\ell)}\circ\cdots\circ \ten L_{\text{\tiny{KANO}}}^{(1)}.
\end{equation}
\begin{equation}\label{eq:kano-layer}
  \ten L_{\text{\tiny{KANO}}}(\mat{a})(\mat{x})
  \;=\;
  \bm{\Phi}_{\theta_1}\!\Bigl(
     \ten F_m^{-1}\!\bigl[\,\boldsymbol{p}_{\theta_2}(\mat{x},\bm{\xi})\,*\,\ten F_m (\mat{a})(\bm{\xi})\,\bigr](\mat{x})\;,\;
     \mat{a}(\mat{x})
  \Bigr),
\end{equation}
where $\bm{\Phi}$ is a KAN sub-network for learnable non-linear activation; $\boldsymbol{p}(\mat{x},\bm{\xi})$ is another KAN sub-network, a pseudo-differential symbol jointly parametrized by both spatial $\mat{x}$ and spectral $\bm{\xi}$ bases\footnote{\citet{pdno} first employed pseudo-differential operator framework for neural operator. They presumed the symbol $\boldsymbol{p}(\mat x,\bm\xi)$ to be separable as $\boldsymbol{p}(\mat x,\bm\xi)=\boldsymbol{p}_{\mat x}(\mat x)\cdot\boldsymbol{p}_{\bm\xi}(\bm\xi)$, and used MLP sub-networks while retaining the lift-up and projection networks of a generic neural operator architecture~\citep{no}.}. Note the ``$\ast$'' notation in $\ten{L}_{\text{\tiny{KANO}}}$ (\ref{eq:kano-layer}) instead of the block-diagonal multiplication notation ``$\cdot$'' in $\ten{L}_{\text{\tiny{FNO}}}$ (\ref{eq:fno_layer}). The spatial basis $\mat{x}$ of the symbol $\boldsymbol{p}(\mat{x},\bm{\xi})$ is convolution (differential) in spectral domain by the dual relationship (\ref{eq:mult_diff_dual}). Therefore, the pseudo-differential symbol calculus of $\boldsymbol{p}(\mat{x},\bm{\xi})$ needs to be done by quantizing on both spatial and spectral domain~\citep{hormanderPDO}, and we choose Kohn-Nirenberg quantization~\citep{kohnNirenberg} to compute the symbol calculus in $\ten{L}_{\text{\tiny{KANO}}}$:

\vspace{-5pt}
\begin{equation}\label{eq:kn_quant}
      \ten F_m^{-1}\!\bigl[\,\boldsymbol{p}(\mat{x},\bm{\xi})\,*\,\ten F_m (\mat{a})(\bm{\xi})\,\bigr](\mat{x})
  :=
  \Bigl(\frac{h}{L}\Bigr)^d
  \sum_{\bm{\xi}\in\Xi} \sum_{\mat{y}\in\mathcal{Y}}
        e^{i(\mat{x}-\mat{y})\cdot\bm{\xi}}\;
        \boldsymbol{p}(\mat{x},\bm{\xi})\,
        \mat{a}(\mat{y}),
\end{equation}
where for a periodic domain $D=(-\frac{L}{2},\frac{L}{2})^d$, $\mathcal{Y}=\{\mat{y}_1,\dots,\mat{y}_m\}\subset D$ is a uniform discretization with spacing $h$ and $\mat{x}\in D$ is an evaluation point. We denote Kohn-Nirenberg quantization (\ref{eq:kn_quant}) as an operator $\mat{Op}_m\!\bigl(\boldsymbol{p}\bigr):=\ten F_m^{-1}\!\bigl[\,\boldsymbol{p}(\mat{x},\bm{\xi})\,*\,\ten F_m \bigr]$ defined by the symbol $\boldsymbol{p}(\mat{x},\bm{\xi})$. In the operator network formulation (\ref{eq:operator_network}) introduced in Section~\ref{sec:fno_background}, KANO's projection $\bm{\Pi}_{\text{\tiny{KANO}}}$ is then defined as:
\begin{equation}\label{eq:kano_proj}
\bm{\Pi}_{\text{\tiny{KANO}}}(\ten G)\;:=\;\mat{Op}_m\!\bigl(\boldsymbol{p}_{\ten G}\bigr),
\qquad
\boldsymbol{p}_{\ten G}\in\arg\min_{\boldsymbol{p}}\;\bigl\|\ten G-\mat{Op}_m(\boldsymbol{p})\bigr\|.
\end{equation}
\paragraph{Symbolic Interpretability of KANO.} By using compact KANs each for the symbol $\boldsymbol{p}(\mat{x},\bm{\xi})$ and non-linear activation $\bm{\Phi}$ in every KANO layer $\ten{L}_{\text{\tiny{KANO}}}$ (\ref{eq:kano-layer}), KANO network $\ten{G}_\theta^{\text{\tiny{KANO}}}$ (\ref{eq:kano-arch}) is fully inspectable by visualizing the learned edges of all its KANs, potentially allowing closed-form symbolic formula of the learned operator with the manual provided by~\citet{kan,kan2}. In addition, recent endeavors have greatly expanded KAN's symbolic recovery capacity to non-smooth, discontinuous targets with high irregularities~\citep{sinckan,rkan,dkan,kanmha,xkan}. All of such advancements are easily and directly applicable in our KANO framework as well, when facing an operator with high irregularity coefficients.
\begin{remark}[\textbf{Complexity analyses of KANO}]
    As apparent in Eq.~(\ref{eq:kn_quant}), KANO layer must perform double sum which can be computationally heavy. However, for the target operator class of variable-coefficient PDEs such as position-dependent dynamics, we show this can be compensated in principle by the parameter efficiency we prove in the following Section~\ref{sec:thm2}. See Appendix~\ref{app:comp_mem_complexity}.
\end{remark}


\subsection{KANO's Dual-Domain Expressivity}
\label{sec:thm2}
\vspace{-5pt}

In contrast to FNO, KANO exploits sparse representations in both the spatial and spectral domains, hence decoupling the scaling of $\epsilon_{\text{proj}}$ and $\epsilon_{\text{net}}$ by never letting the latent map be a dense convolution. For instance, for the quantum harmonic oscillator in Eq.~(\ref{eq:quantum-oscillator}), a KANO layer $\ten{L}_{\text{\tiny{KANO}}}$ (\ref{eq:kano-layer}) can parametrize $\ten{H}$ by taking $\boldsymbol{p}(x,\xi)\approx x^2+\xi^2$, both $-\partial_{xx}$ and $x^2$ terms are each represented where they are sparse, both leveraging the shift form \(\mat{S}_n^{(2)}\) (\ref{eq:shift-x2}). By \emph{jointly} parameterizing the operator in both spatial and frequency domains, KANO \emph{cherry-picks} the sparse representation for every term in position-dependent dynamics, building the right inductive bias well-known to be essential for out-of-distribution generalization and model efficiency~\citep{goyal,trask}.

This dual-domain expressivity of KANO first alleviates the input constraint; we first explain that $\epsilon_{\text{proj}}$ of KANO scales practically by its width $m$ for \emph{any} physical input. Then we provide Theorem~\ref{thm:kano-general}: as long as the KANO projection (\ref{eq:kano_proj}) of an operator generates smooth symbols KAN can easily approximate, $\epsilon_{\text{net}}$ scales practically by compact KAN sub-networks independent of $\epsilon_{\text{proj}}$. In conclusion, \emph{KANO can converge closely as desired to a generic position-dependent dynamics with practical model size using any physical input, robustly generalizing outside the train subspace.}

\vspace{-5pt}
\paragraph{KANO practically has no input constraint.}
According to the quadrature bound from~\citet[Thm.\,1\&2]{dsc}, the error estimate of Kohn-Nirenberg quantization (\ref{eq:kn_quant}) obeys
\begin{equation}\label{eq:knproj_bound}
    \bigl\|\ten{G}-\mat{Op}_{m}(\boldsymbol{p}_{\ten{G}})\bigr\|
          \;\le\;C\,B\,m^{-s},
\end{equation}
given norm-bound (finite-energy)\footnote{Norm here and Equation Eq.~\ref{eq:knproj_bound} is the Sobolev norm} input of $\mathcal A_{B}=\{\mat{u}:\|\mat{u}\|\le B\}$ where $s, C$ are geometric constants. Hence KANO width $m$ scales polynomially by the desired $\epsilon_{\text{proj}}$ given any physical data.
\begin{theorem}[\textbf{KANO stays practical for smooth symbol}]
\label{thm:kano-general}
If the KANO projection of an operator $\ten{G}$, $\bm{\Pi}_{\text{\tiny{KANO}}}(\ten{G})$ (\ref{eq:kano_proj}), generates a finite composition of smooth symbols $\boldsymbol{p}_{\ten{G}}(\mat{x},\bm{\xi})$ and finite-degree non-linearities, the  model size of KANO scales polynomially by the desired accuracy $\varepsilon$.
\vspace{-10pt}
\begin{proof}[Sketch of proof.]
Choosing \(m\!\sim\!(B/\varepsilon)^{1/s}\) scales projection error down to $\varepsilon/2$ by Eq.~(\ref{eq:knproj_bound}). A fixed-width KAN then approximates the symbols to accuracy \(\varepsilon/2\) with \(\mathcal O(\varepsilon^{-d/(2s_p)})\) parameters~\citep[Corol.\,3.4]{expressivitykan} ($s_p$ is a geometric constant).  
The finite-degree non-linearities add only constant-size weights by the activation KAN, so the total parameter count is \(\mathcal O(\varepsilon^{-d/(2s_p)})\). See Appendix~\ref{app:thm2proof} for restatement and full proof.
\end{proof}
\end{theorem}
\begin{corollary}[\textbf{KANO is practical for generic position-dependent dynamics}]
\label{cor:finite-pm}
For a finite composition of spatial and spectral multipliers of maximum $r$-differentiable symbols with finite-degree non-linearity, Theorem~\ref{thm:kano-general} yields
\(
|\Theta|=\mathcal O\bigl(\varepsilon^{-\,d/(2r)}\bigr)
\).
\end{corollary}

\begin{remark}[\textbf{Scope of KANO}]
Recent studies demonstrate that wide lift-up and projection networks are essential for strong performance on high-dimensional benchmarks~\citep{liftup1,liftfup2,liftup3}. In contrast, KANO is designed to prioritize symbolic recovery with robust generalization, complementary to the scope of FNO. Because the core dual-domain expressivity is mathematically agnostic to dimensionality, extending KANO to conventional high-dimensional use cases is a natural and promising direction for future work.
\end{remark}

%% file: results.tex

\vspace{-5pt}
\subsection{Synthetic-Operator Generalization Benchmarks}\label{sec:synth-benchmark}

\begin{table}[t]
\vspace{-15pt}
  \centering
  \caption{Relative \(\ell_2\) losses (\(\times10^{-4}\)) and parameter counts.}
  \label{tab:losstest}
  \footnotesize
  \begin{tabular}{l ccc ccc ccc}
    \toprule
    & \multicolumn{3}{c}{\(\,\ten{G}_1\,\)} 
    & \multicolumn{3}{c}{\(\,\ten{G}_2\,\)}
    & \multicolumn{3}{c}{\(\,\ten{G}_3\,\)} \\
    \cmidrule(lr){2-4}\cmidrule(lr){5-7}\cmidrule(lr){8-10}
    Model (params) 
      & A & B & B/A 
      & A & B & B/A 
      & A & B & B/A \\ \midrule
    FNO (566k)           
      & 6.36  & 98.8  & 15.53 
      & 10.6  & 87.0  & 8.21 
      & 11.4  & 81.4  & 7.14 \\
    U-FNO (579k)        
      & 2.79  & 22.9  & 8.21 
      & 8.14  & 339   & 41.65 
      & 92.4  & 292   & 3.16 \\
    AM-FNO (548k)      
      & 1.08  & 20.9  & 19.35 
      & 1.20  & 16.5  & 13.75 
      & 1.16  & 29.8  & 25.69 \\
    PDNO (538k)         
      & 1.41  & 6.31  & 4.5 
      & 1.92  & 12.1  & 6.3 
      & 4.03  & 27.2  & 6.7 \\
    \textsc{KANO} (152) 
      & 1.04  & 1.44  & 1.38 
      & 0.629 & 0.749 & 1.19 
      & 0.716 & 0.737 & 1.03 \\
    \textsc{KANO\_mlp} (2k) 
      & 3.37  & 6.59  & 1.96 
      & 4.49  & 8.07  & 1.80 
      & 3.59  & 6.87  & 1.91 \\
    \textsc{KANO\_symbolic} 
      & 0.512 & 0.526 & 1.03 
      & 0.498 & 0.500 & 1.00 
      & 0.520 & 0.536 & 1.03 \\
    \bottomrule
  \end{tabular}
  \normalsize
\end{table}

\begin{figure}[t]
\vspace{-15pt}
  \centering
  \includegraphics[width=0.95\textwidth]{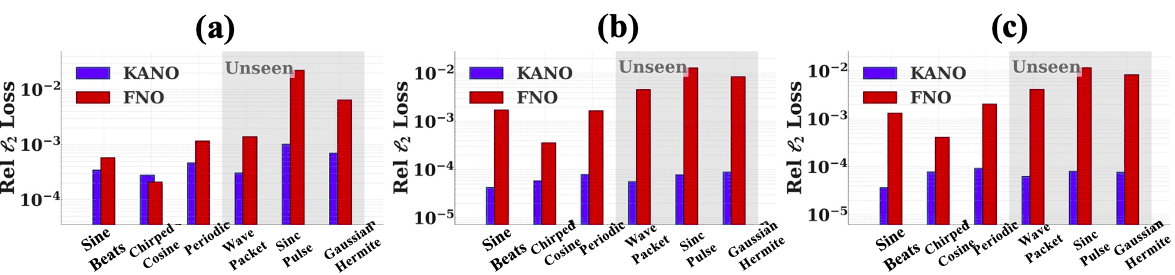}
  \caption{Loss test results. \textbf{(a)} $\ten G_1$ \textbf{(b)} $\ten G_2$ \textbf{(c)} $\ten G_3$. Note the logarithmic scale.}
  \label{fig:loss}
  \vspace{-6pt}
\end{figure}

\begin{figure}[t]
  \centering  \includegraphics[width=0.95\textwidth]{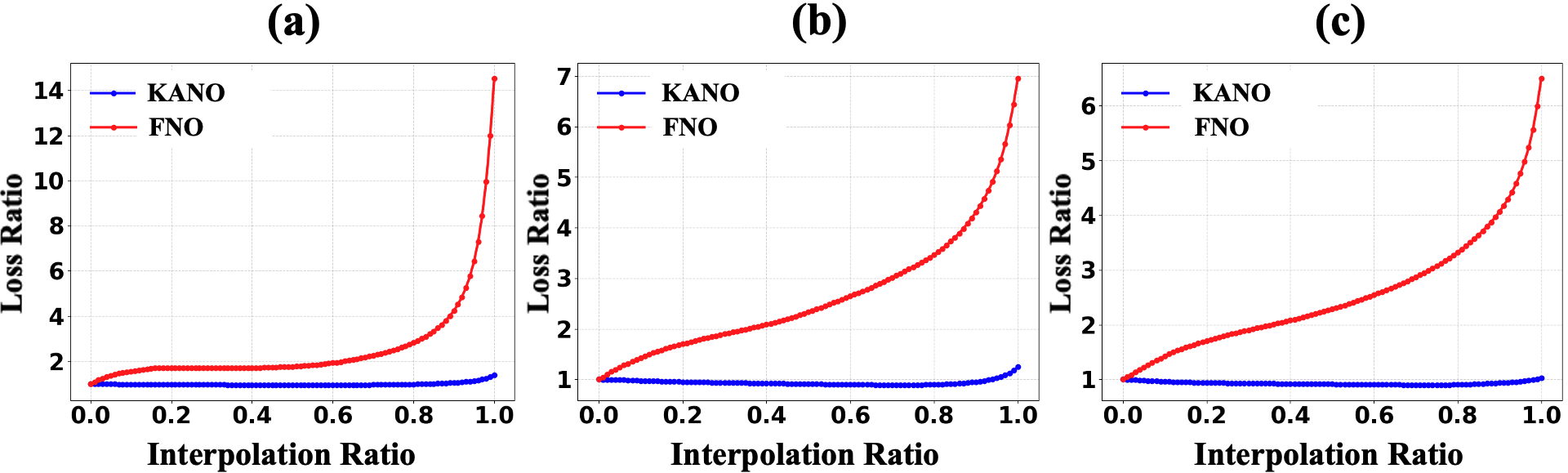}
  \caption{Interpolation test results. \textbf{(a)} $\ten G_1$ \textbf{(b)} $\ten G_2$ \textbf{(c)} $\ten G_3$. }
  \label{fig:interpolation}
  \vspace{-6pt}
\end{figure}
\begin{figure}[t]
  \centering
  \includegraphics[width=\textwidth]{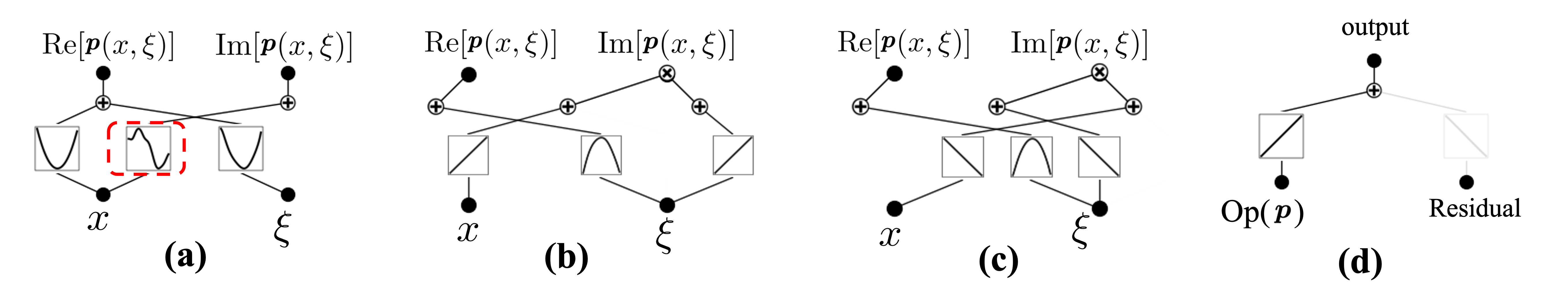}
\vspace{-12pt}
  \caption{\textbf{(a)} $\boldsymbol{p}(x,\xi)$ of $\ten{G}_1$. The middle edge does not contribute to the output. \textbf{(b)} $\boldsymbol{p}(x,\xi)$ of $\ten{G}_2$. \textbf{(c)} $\boldsymbol{p}(x,\xi)$ of $\ten{G}_3$. \textbf{(d)} $\bm{\Phi}$ of $\ten{G}_3$. Edge of the residual in \textbf{(d)} looks linear, so we compared two scenarios, linear and cubic, which the latter achieved lower loss and better generalization.}
  \label{fig:synspline}
  \vspace{-12pt}
\end{figure}

We benchmark FNO-based models and KANO on three position-dependent operators:
$$
\begin{aligned}
  \ten{G}_1f &= x^{2}\cdot f \;-\; \partial_{xx}f, &
  \ten{G}_2f &= x\cdot\partial_xf \;+\; \partial_{xx}f, &
  \ten{G}_3f &= f^{3} + x\cdot\partial_xf + \partial_{xx}f .
\end{aligned}
$$

\vspace{-5pt}
Our goal is to quantify and compare the generalization of each model. We train the models \emph{only} with Group A dataset and evaluate them on the \emph{unseen} Group B dataset.
\begin{itemize}[left=0cm]
\vspace{-5pt}
\item \textbf{Group A (Training families):} Periodic, Chirped Cosine, Sine Beats. 
\item \textbf{Group B (Testing families):} Sinc Pulse, Gaussian$\times$Hermite, Wave Packet.
\end{itemize}

\vspace{-5pt}
 For each operator, we generate \(2000\) train pairs from Group A and \(400\) test pairs from Group B to evaluate the generalization by comparing the ratio between the average relative $\ell_2$ loss over each group (\textbf{Loss Test}). We also interpolate the Group A and B function samples in 100 steps, apply ground-truth operators in each step to build the interpolated dataset, and evaluate the loss ratio to that of the Group A samples (\textbf{Interpolation Test}). We trained FNO models of 2 layers, 64 width with no mode truncation, and used one-layer KANs of grid 10 cubic B-splines edges for the KANO model. For other FNO variant baselines, we used 2-layer models of similar size to FNO. Lastly, we trained a KANO variant for an ablation study, which we replaced the KAN subnetworks with compact MLPs of 32 hidden width and 2 hidden layers (KANO\_MLP). See Appendix~\ref{app:syntheticdata} for experiment details.  We used Adam optimizer and relative $\ell_2$ loss for training.

\paragraph{Results.} As shown in Table~\ref{tab:losstest} and Figure~\ref{fig:loss}, KANO shows consistent losses over Group A and Group B, validating its robust generalization ability for position-dependent operators. KANO\_MLP also shows comparable out-of-distribution performance, which suggests that the generalization ability of KANO stems from its dual-domain architecture apart from KAN. 
In contrast, FNO shows fragile out-of-distribution behavior on Group B dataset with the significant loss increases. U-FNO~\citep{ufno} and AM-FNO~\citep{amfno} show even worse results. On the other hand, PDNO~\citep{pdno} shows the most stable generalization among FNO families, although not as robust as KANO and KANO\_MLP. Along with the ablation study, this confirms that the pseudo-differential operator framework is judicious for robust generalization on position-dependent operators, while primarily relying on only spectral kernels makes the model fragile out of train distribution even with localized enhancements. Interpolation test results in Figure~\ref{fig:interpolation} further empirically validate our theories. The FNO curves (red) of Figure~\ref{fig:interpolation} show slow increases on early and mid-interpolation, suggesting that the FNO’s learned in-sample mappings are yet close to the ground-truth operators. However, the FNO curves abruptly soar up in the latter ratio, suggesting that the interpolated functions are now far outside the train distribution. These results, together with KANO’s one-order–of–magnitude lower loss at just $0.03\%$ of FNO’s size, are consistent with our claims in Theorem~\ref{thm:fno-fail} and Theorem~\ref{thm:kano-general}. 

After convergence, we visualized the embedded KANs (Figure~\ref{fig:synspline}). We then froze these learned symbols and continued training, referring to this variant as KANO\_symbolic. KANO\_symbolic recovered the exact symbolic coefficients of the ground-truth operator to within the fourth decimal place (Table~\ref{tab:evidence}). KANO’s loss matches KANO\_symbolic’s loss in Table~\ref{tab:losstest}, confirming that KANO converged close to the ground-truth operator.

\begin{table}[h]
\vspace{-15pt}
  \centering
  \caption{Ground-truth vs.\ learned operators (coefficients rounded to $4^{th}$ decimal place).}
  \label{tab:evidence}
  \footnotesize
  \begin{tabular}{@{}l @{\hspace{0.5cm}} p{7.5cm}@{}}
    \toprule
    \bf Ground-truth operator & \bf Learned KANO operator \\ \midrule
    \(\ten{G}_1f = x^{2}\cdot f - \partial_{xx}f\) &
      \(\tilde{\ten{G}}_1f = (x^{2}+0.0003)\cdot f - \partial_{xx}f\) \\
    \(\ten{G}_2f = x\cdot\partial_x f + \partial_{xx}f\) &
      \(\tilde{\ten{G}}_2f = 0.9996\,x\cdot\partial_x f + \partial_{xx}f - 0.0003\,f\) \\
    \(\ten{G}_3f = f^{3} + x\cdot\partial_x f + \partial_{xx}f\) &
      \makecell[l]{%
        \(\tilde{\ten{G}}_3f = 1.0001\,f^{3}
                      + 0.99997\,x\cdot\partial_x f
                      + 0.99997\,\partial_{xx}f\)\\
        \hphantom{\(\tilde{\ten{G}}_3f =\)}\(-\,0.0002\,f^{2}
                      - 0.0003\,f
                      - 0.0001\)} \\ \bottomrule
  \end{tabular} \normalsize
\end{table}

\subsection{Long-Horizon Quantum Dynamics Benchmark}\label{sec:qkano-setting}


We provide this benchmark on two position-dependent quantum dynamics: the quartic double‐well Hamiltonian (DW) and the nonlinear Schrödinger equation with cubic nonlinearity (NLSE):
\[
i\partial_t\psi \;=\; -\tfrac12\partial_{xx}\psi + w(x)\cdot\psi \,\,(\text{DW}),
\quad
i\partial_t\psi \;=\; -\tfrac12\partial_{xx}\psi + w(x)\cdot\psi + |\psi|^{2}\cdot\psi \,\,(\text{NLSE}),
\]
where $w(x)=x^{4}-\bigl(x-\tfrac1{32}\bigr)^{2}+0.295$. 

We generate $200$ initial states and yield the state trajectories by the Hamiltonians, sampling momentum/position probability mass functions (PMFs) every $0.1\text{ms}$ for $100$ time steps. The first $10$ time steps are used for training, and the rest are used to evaluate the long-horizon prediction.


We modify KANO to capture the quantum state evolution: \textbf{\emph{Q-KANO}}. Symbol $\boldsymbol{p}_{\theta}$ is parametrized as $\exp\!\bigl[-i\Delta T\,\bm{\phi}_\theta(\mat{x},\bm{\xi})\bigr]$, where $\Delta T=0.1\,\text{ms}$.  The adaptive activation is also defined as a complex exponential with learned phase \(\bm{\vartheta} = \bm{\Phi}_{\theta}(\lvert \mat{Op}_m(\boldsymbol{p}_\theta)\bm{\psi}\rvert, \angle \mat{Op}_m(\boldsymbol{p}_\theta)\bm{\psi})\) for input wave function $\bm{\psi} (\mat{x})$: 
  \begin{equation}
      \ten{G}^{\tiny{\text{Q-KANO}}}_\theta[\bm{\psi}] =\mat{Op}_m(\exp\!\bigl[-i\Delta T\,\bm{\phi}_\theta(\mat{x},\bm{\xi})\bigr])\bm{\psi} \;\cdot\; e^{-i\Delta T\,\bm{\vartheta}}.
  \end{equation}
We investigate three supervision scenarios: \textbf{Full}-type training with full wave function, idealistic yet physically unattainable, \textbf{Pos}-type training with only position PMF, physically realistic yet the least informative, and \textbf{pos\&mom}-type training with both position \emph{and} momentum PMFs, which remains physically attainable while providing richer information although not full. We use Adam optimizer for all trainings. See Appendix~\ref{app:quantumbenchmark} for experiment details.

\begin{figure}[t]
  \centering
  \includegraphics[width=0.9\textwidth]{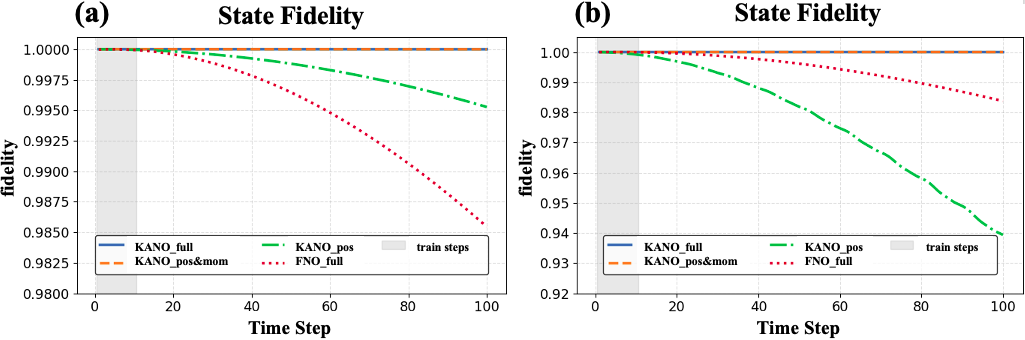}
  \caption{State fidelity over 100 time steps. \textbf{(a)} DW  \textbf{(b)} NLSE}
  \label{fig:fidelitydrop}
  \vspace{-5pt}
\end{figure}

\begin{wraptable}{r}{4in}
  \centering
  \vspace{-20pt}
  \caption{State infidelity after 90 additional time-evolution steps.}
  \label{tab:fidelity-100}
  \footnotesize
  \begin{tabular}{lcc}
    \toprule
    \multirow{2}{*}{Model \& Train Type} &
    \multicolumn{2}{c}{State Infidelity} \\
    \cmidrule(l){2-3}
    & Double-Well & NLSE \\
    \midrule
    \textsc{FNO} (full)                     & \(1.5\times10^{-2}\)  & \(1.6\times10^{-2}\)  \\
    \textsc{Q-KANO} (full)                  & \(6.3\times10^{-6}\)  & \(6.8\times10^{-6}\)  \\
    \textsc{Q-KANO} (pos \& mom)            & \(6.3\times10^{-6}\)  & \(5.6\times10^{-6}\)  \\
    \textsc{Q-KANO} (pos)                   & \(4.7\times10^{-3}\)  & \(6.1\times10^{-2}\)  \\
    \addlinespace
    \textsc{Q-KANO\_mlp} (pos \& mom)            & \(7.7\times10^{-6}\)  & \(8.5\times10^{-6}\)  \\
    \addlinespace
    \textsc{Q-KANO\_symbolic} (full)        & \(2.0\times10^{-8}\)  & \(2.0\times10^{-8}\)  \\
    \textsc{Q-KANO\_symbolic} (pos \& mom)  & \(2.0\times10^{-8}\)  & \(3.0\times10^{-8}\)  \\
    \textsc{Q-KANO\_symbolic} (pos)         & \(5.3\times10^{-2}\)  & \(6.1\times10^{-2}\)  \\
    \bottomrule
  \end{tabular}\normalsize
\end{wraptable}
\paragraph{Results.} We evaluate state infidelity\footnote{For predicted state $\tilde{\bm{\varphi}}$ and ground-truth state $\bm{\varphi}$, the state fidelity $F$ is defined as the inner product between them ($F:=<\tilde{\bm{\varphi}},\bm{\varphi}>$), and the state infidelity is defined as $(1-F)$, hence shows how distant two states are.} between ground-truth evolution and model prediction at each time step (Table~\ref{tab:fidelity-100}, Figure~\ref{fig:fidelitydrop}). In case of KANO, the \textbf{pos\,\&\,mom}-type training achieves indistinguishable infidelity from the ideal \textbf{full}-type training baseline. The ablation study with MLP variant of Q-KANO achieved comparably low state infidelity by \textbf{pos\,\&\,mom}-type training as well. Meanwhile, the \textbf{pos-type} training displays a clear increase in infidelity, especially on the NLSE.

In contrast, even with \textbf{full} type training, FNO fails to maintain low state infidelity after the long-horizon propagation as expected. Iterative time evolution pushes the wave function far outside the train convex hull, and FNO’s learned in‐sample mapping deviates from the ground\mbox{-}truth evolution rapidly, leading to four orders of infidelity increase compared to KANO.

\begin{figure}[t]
  \centering
  \includegraphics[width=0.9\textwidth]{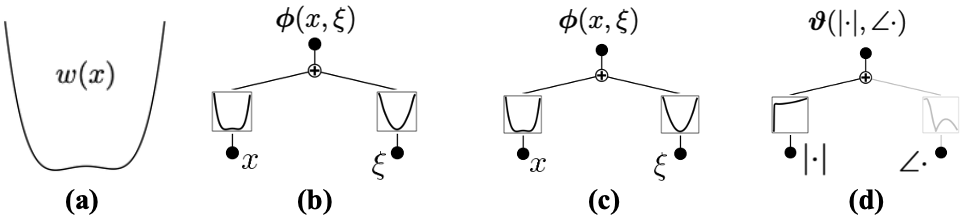}
  \caption{\textbf{pos\&mom} type training results. \textbf{(a)} Structure of the potential $w(x)$ \textbf{(b)} $\boldsymbol{p}(x,\xi)$ of DW. \textbf{(c)} $\boldsymbol{p}(x,\xi)$ of NLSE. \textbf{(d)}  $\varphi(\lvert\cdot\rvert,\angle\cdot)$ of NLSE. Potential $w(x)$ structure is clearly reconstructed.}
  \label{fig:quantumspline}
  \vspace{-10pt}
\end{figure}

Table~\ref{tab:dw-nlse-operators} juxtaposes the learned symbols with that of the ground‐truth Hamiltonians and Figure~\ref{fig:quantumspline} shows the KAN visualizations from \textbf{pos\,\&\,mom}-type training. With \textbf{full}-type training coefficients are recovered to the fourth decimal place, vindicating the ideal capacity of KANO when the information is fully provided. Under the realistic \textbf{pos\,\&\,mom}-type training, the reconstruction remains accurate except for two terms: the constant (global phase) and the NLSE’s cubic coefficient. Both discrepancies are predicted by quantum observability: global phases cancel in all PMFs, and the Kerr coefficient enters only through higher-order correlations that become harder to estimate from finite-shot statistics. Q-KANO faithfully reveals what the data support and nothing more.

\begin{table}[t]
\vspace{-10pt}
  \centering
  \caption{Ground truth vs.\ learned symbols.  Coefficients rounded to $4^{th}$ decimal place.}
  \label{tab:dw-nlse-operators}
  \footnotesize
  \begin{tabular}{@{}l@{\quad}c@{\quad}p{9.0cm}@{}}
    \toprule
    Hamiltonian & Train Type & Learned symbolic structure \\
    \midrule
    \rowcolor{gray!15}
      \multirow{3}{*}{\cellcolor{white}\textbf{DW}} 
      & ground truth &
        $x^{4}-x^{2}+0.0625\,x+0.295+0.5\,\xi^{2}$ \\ \cmidrule(lr){2-3}
      & full &
        $1.0004\,x^{4}+0.0001\,x^{3}-1.0013\,x^{2}+0.0625\,x+0.2955+0.5\,\xi^{2}$ \\ \cmidrule(lr){2-3}
      & pos \& mom &
        $1.0003\,x^{4}+0.0001\,x^{3}-1.0008\,x^{2}+0.0623\,x+0.0001+0.5\,\xi^{2}$ \\
    \midrule
    \rowcolor{gray!15}
      \multirow{3}{*}{\cellcolor{white}\textbf{NLSE}}
      & ground truth &
        $x^{4}-x^{2}+0.0625\,x+0.295+0.5\,\xi^{2}+|\psi|^{2}$ \\ \cmidrule(lr){2-3}
      & full &
        $1.0005\,x^{4}-0.0001\,x^{3}-1.0014\,x^{2}+0.0626\,x+0.2942+0.5\,\xi^{2}+0.9815|\psi|^{2}+0.0110|\psi|$ \\ \cmidrule(lr){2-3}
      & pos \& mom &
        $0.9999\,x^{4}-0.0003\,x^{3}-1.0001\,x^{2}+0.0630\,x+0.1141+0.5\,\xi^{2}+0.9514|\psi|^{2}-0.5504|\psi|$ \\
    \bottomrule
  \end{tabular} \normalsize
\end{table}

%% file: discussion.tex
\vspace{-5pt}
We have presented the Kolmogorov–Arnold Neural Operator, an interpretable neural operator expressive on a generic position-dependent dynamics. KANO cherry-picks sparse representations of each term via jointly parametrizing on both spectral and spatial bases, and achieves robust generalization outside train distribution while exposing clear tractable representation via its KAN sub-networks. In all our benchmarks, KANO has successfully recovered the ground-truth operators accurately to the fourth decimal place in coefficients. In addition to the superior out-of-distribution generalization, KANO has also achieved orders of magnitude lower losses with less than $0.03\%$ of the model size compared to the FNO baseline. KANO shifts operator learning from an opaque, surrogate-based paradigm towards interpretable data-driven scientific modeling, and provides robust empirical evidence supporting its enhanced dual-domain expressivity and interpretability.

%% file: reproducibility.tex
 All code, configurations, and dataset artifacts required to reproduce the experiments are provided in \url{https://github.com/jinucsb/KANO}, along with step-by-step instructions for rerunning training and evaluation. Details in the data generation and implementation are provided in Appendix~\ref{app:impl}, and the full proofs of Theorems and Lemma are provided in Appendix~\ref{app:proofs} along with their mathematical restatements.

%% file: ethics.tex
All authors of this work sincerely adhered to the ICLR Code of Ethics. We do not expect any potential violation to the best of our knowledge.

%% file: ack.tex
We thank Nina Miolane, Adele Myers, Louisa Cornelis for helpful early discussions and Anima Anandkumar for her feedback on our paper draft. Jin Lee and Zheng Zhang were supported by NSF ECCS-2328281. Murphy Yuezhen Niu was supported by the U.S. National Science Foundation grant CCF-2441912 (NSF CAREER), Air Force Office of Scientific Research under award number FA9550-25-1-0146, and   the U.S. Department of Energy,  Office of  Advanced Scientific Computing Research  under Award Number DE-SC0025430. 

%% file: appendix_notation.tex
\begin{table}[htbp]
\centering\small
\renewcommand{\arraystretch}{1.15}
\setlength{\tabcolsep}{8pt}
\caption{Main symbols and notation used in the paper.\vspace{4pt}}
\begin{tabular}{p{0.30\textwidth} p{0.62\textwidth}}
\toprule
\textbf{Symbol} & \textbf{Meaning / Definition}\\
\midrule
\multicolumn{2}{l}{\emph{Domains, spaces, and operators}}\\

$D:=(-L/2,L/2)^{d}$ & Periodic $d$-dimensional spatial box of side length $L$\\
$\ten{G}$ & Ground-truth solution operator to be learned\\[2pt]

\multicolumn{2}{l}{\emph{Spectral \& spatial sampling}}\\
$\Xi=\{\bm{\xi}_{1},\ldots,\bm{\xi}_{m}\}$ & Retained Fourier modes (truncated spectrum); $m=|\Xi|$\\
$\ten{F}$ & Fourier transform\\
$\ten{F}_{m},\;\ten{F}_m^{-1}$ & Truncated Fourier transform by $\Xi$ and its band-limited inverse\\
$\mathcal{Y}=\{\mat{y}_{1},\ldots,\mat{y}_{m}\}\subset D$ & Uniform spatial grid\\
$h$ & Grid spacing of $\mathcal Y$\\[2pt]

\multicolumn{2}{l}{\emph{Fourier Neural Operator (FNO)}}\\
$\ten{L}_{\tiny{\text{FNO}}}$ & Single FNO layer\\
$\ten{G}^{\tiny{\text{FNO}}}_{\theta}$ & FNO network\\
$\mat{R}_{\theta}(\bm{\xi})$ & Learnable block-diagonal spectral multiplier\\
$\mat{W}_{\theta}$ & Point-wise learnable linear map\\
$\sigma(\cdot)$ & Point-wise non-linear activation\\[2pt]

\multicolumn{2}{l}{\emph{KANO layer, symbol calculus, and projection}}\\
$\ten{L}_{\tiny{\text{KANO}}}$ & Single KANO layer\\
$\boldsymbol{p}(\mat{x},\bm{\xi})$ & Learnable pseudo-differential symbol\\
$\mat{Op}_{m}(\boldsymbol{p})$ & Kohn-Nirenberg quantization of width $m$ defined by $\boldsymbol{p}$\\
$\bm{\Pi}_{\tiny{\text{KANO}}}(\ten{G})$ & KANO projection of $\ten{G}$\\
$\ten{G}^{\tiny{\text{KANO}}}_{\theta}$ & KANO network \\\
$\bm{\Phi}_{\theta}$ & Learnable activation\\[2pt]

\multicolumn{2}{l}{\emph{Kolmogorov--Arnold Network (KAN) primitives}}\\
$\phi^{(\ell)}_{q,p}(\cdot)$ & 1D edge function on layer $\ell$, connecting $p^{\text{th}}$ node of layer $\ell$ to $q^{\text{th}}$ node of layer $(\ell+1)$\\
$b(t),\;\{B_{i}(t)\}$ & Base function and B-spline basis used to parametrize $\phi^{(\ell)}_{q,p}(t)$\\[2pt]

\multicolumn{2}{l}{\emph{Q-KANO (quantum dynamics) notation}}\\
$\psi(x)$ & Input wave function\\
$w(x)$ & Quartic double-well potential\\
$\Delta T$ & Time step of propagation \\
$\bm{\phi}_\theta(\mat{x},\bm{\xi})$ & Parametrized phase for symbol $\boldsymbol{p}(\mat{x},\bm{\xi})$ of Q-KANO\\
$\bm{\vartheta}_\theta(\lvert\cdot\rvert,\angle\cdot)$ & Parametrized phase for non-linear activation of Q-KANO\\
$\ten{G}^{\tiny{\text{Q-KANO}}}_\theta$ & Q-KANO network\\

\multicolumn{2}{l}{\emph{Function spaces}}\\
$L^{2}(D)$ & Square-integrable function space on domain $D$\\
$H^{s}(D)$ & Sobolev function space of order $s\ge 0$ on domain $D$\\

\bottomrule
\end{tabular}
\label{tab:notation}
\end{table}

%% file: appendix_impl.tex
\subsection{Synthetic Operator Benchmark}
\label{app:syntheticdata}

All experiments are carried out on periodic functions  
\(f:\mathbb T\to\mathbb R\) with  
\(\mathbb T=(-\pi,\pi]\) and a uniform trigonometric grid  
\[
   x_{j}=x_{\min}+j\,\Delta x, 
   \quad
   \Delta x = \frac{2\pi}{N}, 
   \quad
   j=0,\dots,N-1 ,
\]
with \(N=128\).  Unless noted otherwise every random quantity is drawn
\emph{independently for every sample}.

\paragraph{Outer envelope.}
To avoid the Gibbs phenomenon all basis functions are multiplied by a
smooth taper that decays to zero in a \( \nicefrac{\pi}{6}\)-wide buffer
near the periodic boundary:
\[
A(x)=
\begin{cases}\label{eq:envelope}
  1, & |x|\le 5\pi/6,\\[4pt]
  \cos^{4}\!\Bigl[
      \dfrac{|x|-5\pi/6}{\pi/6}\,
      \dfrac{\pi}{2}
     \Bigr], & 5\pi/6<|x|<\pi,\\[6pt]
  0, & |x|\ge \pi.
\end{cases}
\]
The full “base” function is always  
\(f_{\text{base}}(x)=A(x)\,g(x)\).

\paragraph{Spectral derivatives and ground-truth operator.}
Derivatives are computed with an exact Fourier stencil:
\[
  f'(x) \;=\; \ten{F}^{-1}\!\bigl[\,i\xi\,\widehat f(\xi)\bigr],
  \qquad
  f''(x)\;=\; \ten{F}^{-1}\!\bigl[-\xi^{2}\widehat f(\xi)\bigr].
\]

$U[a,b]$ denotes random digit drawn from range $[a,b]$.

\paragraph{Training families (\textbf{Group A})}
\begin{enumerate}[leftmargin=2.2em,label=\textbf{A\arabic*.}]
\item \textbf{sine\_beats}:  
      \[
        g(x)=\sin(\omega_{1}x+\phi_{1})\,
              \sin(\omega_{2}x+\phi_{2}),
        \quad
        \omega_{i}=8\,U[0.5,3],\;
        \phi_{i}=U[0,2\pi].
      \]

\item \textbf{chirped\_cosine}:  
      \[
        g(x)=\cos\!\bigl(\alpha x^{2}\bigr), 
        \qquad 
        \alpha =12\,U[0.5,2].
      \]

\item \textbf{periodic} (random harmonic series):  
      \[
        g(x)=\sin(\omega x+\phi_{1})+\cos(\omega x+\phi_{2}),
        \qquad
        \omega=8\,U[0.5,3],\;
        \phi_{1,2}=U[0,2\pi].
      \]
\end{enumerate}

\paragraph{Unseen families (\textbf{Group B})}
\begin{enumerate}[leftmargin=2.2em,label=\textbf{B\arabic*.}]
\item \textbf{wave\_packet}:  
      \[
        g(x)=
        \exp\!\Bigl[-\tfrac{(x-\mu)^{2}}{2\sigma^{2}}\Bigr]\,
        \sin\bigl(\omega x+\phi\bigr),
        \;
        \mu=U[-2,2],\;
        \sigma=\tfrac{1}{12}\,U[0.5,2],\;
        \omega = 12\,U[2,6],\;
        \phi=U[0,2\pi].
      \]

\item \textbf{sinc\_pulse}:  
      \[
        g(x)=
        \begin{cases}
           \dfrac{\sin(\alpha x)}{\alpha x}, & |x|>10^{-12},\\[6pt]
           1, & |x|\le 10^{-12},
        \end{cases}
        \qquad
        \alpha = 12\,U[0.5,3].
      \]

\item \textbf{gaussian\_hermite}:  
      \[
        g(x)=H_{n}\!\Bigl(\dfrac{x-\mu}{\sigma}\Bigr)\,
              \exp\!\Bigl[-\tfrac{(x-\mu)^{2}}{2\sigma^{2}}\Bigr],
        \;
        n\in\{1,2,3\} \text{ uniform},\;
        \mu=U[-2,2],\;
        \sigma=\tfrac{1}{8}\,U[0.5,2],
      \]
      where \(H_{n}\) is the degree-\(n\) Hermite
      polynomial.
\end{enumerate}

\vspace{4pt}
\subsubsection*{Normalization}
Each realization is divided by its maximum absolute value,
\(\|f\|_{\infty}\), to obtain  
\(\|f\|_{\infty}=1\).  The envelope guarantees
periodicity and keeps the numerical spectrum sharply band-limited.

\paragraph{Sample counts.}
\(\;\#\text{train}=2000\) samples from the three Group A families for train data and
\(\;\#\text{test}=400\) samples each from the Group A and Group B families for generalization tests.

\subsection{Quantum Dynamics Benchmark}\label{app:quantumbenchmark}

We model a quantum apparatus with $200$ state-preparation protocols each with perfect reproducibility, capable of generating an identical initial state whose wave function is drawn from one of the three families: \emph{Periodic}, \emph{Gaussian wave-packet}, and \emph{Gaussian$\times$Hermite}. The prepared initial states evolve under one of two unknown, time-independent Hamiltonians, and two arrays of $128$ detectors measure position and momentum on uniform grids, yielding probability mass functions (PMFs) every $0.1\text{ms}$ for $100$ time steps. PMFs collected from the first $10$ time steps are used for training the models, and the rest of the PMFs collected from the remaining 90 time steps are used to evaluate the long-horizon fidelity drop beyond the train steps.

\subsubsection{Quantum Apparatus Assumptions}
\begin{enumerate}[left=0cm]
  \item \textbf{State preparation.}  A \emph{collection of calibrated protocols}
        can each prepare a designated initial wave-function each of one of three real-valued families  
        \emph{Periodic}, \emph{Gaussian wave-packet}, or
        \emph{Gaussian–Hermite}:
        $\bm{\psi}^{(m)}_{0}(x)\!\in\!L^{2}(\mathbb T)$\footnote{Square-integrable function space.}, $m=1,\dots,200$.
        Repeated shots under the \emph{same} protocol start from \textbf{exactly}
        the same $\bm{\psi}^{(m)}_{0}$, enabling trajectory-level reproducibility for
        every member of the ensemble.
  \item \textbf{Hamiltonian stability.}  The (unknown) Hamiltonian is \emph{time–independent}, so
        trajectories are perfectly repeatable once $\bm{\psi}_0$ is fixed.
  \item \textbf{Dual-basis detection.}  Two $128$-grid projective detectors measure the
        position basis $\{|x_i\rangle\}$ and the momentum basis
        $\{|\xi_j\rangle\}$, yielding empirical probability mass functions (PMFs)
        $\hat p_x(i)=|\psi(x_i)|^2$ and
        $\hat p_{\xi}(j)=|\widehat\psi(\xi_j)|^2$ on a common torus grid
        $\mathbb T_L,\,L=4$.
\end{enumerate}

\subsubsection{Data Generation Details}
For each of $200$ distinct sample trajectories we
\begin{enumerate}[left=0cm]
  \item draw the initial wave function and propagate on the Hamiltonian with a high-resolution Strang split:
        $\delta t=1\,\mu\text{s}$ for $10\,000$ micro-steps, producing
        coarse snapshots every $100$ steps
        ($\Delta T=0.1\,\text{ms}$, $T=1,\dots,100$);
  \item store $(\bm{\psi}_T,\;\mat{p}_x^T,\;\mat{p}_{\xi}^T)$ where
        $p_x^T(i)=|\psi_T(x_i)|^2$ and
        $p_{\xi}^T(j)=|\widehat\psi_T(\xi_j)|^2$.
\end{enumerate}
Only the first $10$ coarse steps are used for training; the remaining $90$
steps test fidelity drop on long-horizon. All simulations employ an
$n=128$-point FFT grid to match the detectors.

\paragraph{Spatial discretization.}
We place the problem on a periodic box of length \(L=4\) with
\(N=128\) grid points
\(
  x_{j}=x_{\min}+j\Delta x,\;
  \Delta x=L/N.
\)
Periods suppress wrap–around artifacts because every initial state is
tapered by the smooth envelope
\(
  A(x)\) defined in Appendix~\ref{app:syntheticdata}.  Spatial derivatives are taken spectrally:
let \(\xi_{m}=2\pi m/L\) for \(m=-N/2,\dots,N/2-1\).  Writing
\(\widehat\psi_{m}=\ten F[\psi](\xi_{m})\),
\[
  \partial_{x }\psi
     =\ten F^{-1}\!\bigl[i\,\xi_{m}\widehat\psi_{m}\bigr],
  \qquad
  \partial_{xx}\psi
     =\ten F^{-1}\!\bigl[-\xi_{m}^{2}\widehat\psi_{m}\bigr].
\]

\paragraph{Strang–splitting time integrator.}
Let \(\ten{K}:=-\tfrac12\partial_{xx}\) (kinetic),
\(\ten{V}:\psi\mapsto w(x)\cdot\psi\) (potential) and
\(\ten{N}:\psi\mapsto|\psi|^{2}\cdot\psi\) (cubic nonlinearity).  
With time step \(\Delta t\) the second–order Strang factorization reads
\[
  e^{(\ten{K}+\ten{V}+\ten{N})\Delta t}
  \;=\;
  e^{\tfrac{\Delta t}{2}(\ten{K}+\ten{V})}\;
  e^{\Delta t \ten{N}}\;
  e^{\tfrac{\Delta t}{2}(\ten{K}+\ten{V})}
  \;+\;\mathcal O(\Delta t^{3}).
\]
Because \(\ten K\) is diagonal in Fourier space and \(\ten V\) in real space we
implement each half step explicitly:
\[
\boxed{\;
\begin{aligned}
   \psi          &\;\xleftarrow{\;\;\ten{V}/2\;\;}\;
                   e^{-\,\tfrac{i}{2}\Delta t\,w(x)}\psi,\\[4pt]
   \widehat\psi  &\;\xleftarrow{\;\;\ten{K}/2\;\;}\;
                   \ten F[\psi],\quad
                   \widehat\psi_{m}\gets
                   e^{-\,\tfrac{i}{4}\Delta t\,\xi_{m}^{2}}\widehat\psi_{m},\\[4pt]
   \psi          &\;\xleftarrow{\;\;\;\ten N\;\;\;}\;
                   \ten F^{-1}[\widehat\psi],\quad
                   \psi\gets e^{-\,i\Delta t|\psi|^{2}}\psi,\\[4pt]
   \text{repeat }\ten{K}/2\text{ and }\ten{V}/2.
\end{aligned}}
\]
The $L^{2}$ norm is renormalized after every macro
step to compensate numerical drift. For DW Hamiltonian, non-linearity time evolution is omitted.

\paragraph{Integrator parameters.}
We use an inner step \(\delta t=10^{-6}\).  A \textit{macro} step of
\(\Delta T=100\,\delta t=10^{-4}\) is recorded
and the sequence is propagated for
\(T_{\max}=100\,\Delta T=10^{-2}\).

\paragraph{Initial–state families (real wave‐functions).}
Each sample draws a real profile \(f(x)\) from \(\{A_{1},A_{2},A_{3}\}\)
below, multiplies it by the envelope \(A\) and normalizes it in
\(L^{2}\):
\[
  \psi_{0}(x)=\frac{A(x)\,f(x)}{\lVert A f\rVert_{2}}.
\]
\begin{enumerate}[leftmargin=2.4em,label=\textbf{A\arabic*.}]
\item \textbf{random harmonic series}
      \[
        f(x)=\sum_{m=0}^{5}
              \tfrac12\,\xi_{m}^{(c)}\cos(2\pi m\xi)
             +\tfrac12\,\xi_{m}^{(s)}\sin(2\pi m\xi),
        \quad
        \xi=\frac{x+2}{4},\;
        \xi_{m}^{(\cdot)}\sim\mathcal N(0,1).
      \]
\item \textbf{Gaussian wave packet}  
      \(f(x)=\exp[-(x-x_{0})^{2}/(2\sigma^{2})]\)
      with \(x_{0}\!\sim\!U[-0.4\pi,0.4\pi]\) and
      \(\sigma\!\sim\!U[0.1,0.3]\).
\item \textbf{Gaussian–Hermite mode}
      \(
        f(x)=H_{n}\bigl(\tfrac{x-x_{0}}{\sigma}\bigr)
              \exp[-(x-x_{0})^{2}/(2\sigma^{2})]
      \)
      with
      \(n\in\{0,1,2\}\) uniform,
      \(x_{0},\sigma\) as above,
      and \(H_{n}\) the Hermite polynomial.
\end{enumerate}

\paragraph{Stored quantities.}
For every sample index \(s\) and every snapshot \(T\in\{0,1,\dots,100\}\)
we save
\[
  \bigl(\mat{x},\;
        \bm{\psi}(x,T),\;
        |\bm{\psi}(x,T)|^{2},\;
        |\widehat{\bm{\psi}}(\xi,T)|^{2}\bigr)
  \quad\longrightarrow\quad
  \text{\tt wavefunc},\; \text{\tt pos\_pdf},\; \text{\tt mom\_pdf}.
\]
All arrays are written in {\small\texttt{float32}} except the complex
wave‐function, stored as {\small\texttt{complex64}}.  Altogether one
call to
\texttt{generate\_dataset(num\_samples{=}200)} produces
\(200\times101\times4=80{,}800\) labeled records.

\subsubsection{Train Type Details}
Let $\bm{\psi}_\Theta^{(T)}$ be the $T$-step prediction of Q-KANO given
$\bm{\psi}_0$.  
\begin{center}
\begin{tabular}{lll}
\toprule
\textbf{Train Type} & \textbf{Train Dataset} & \textbf{Loss function} \\ \midrule
\textbf{full}      & complex $\bm{\psi}^{(T)}$ & 
$\displaystyle
  \mathcal L_{full}
  =\frac{\|\bm{\psi}_\Theta^{(T)}-\bm{\psi}^{(T)}\|_2}{\|\bm{\psi}^{(T)}\|_2}$ \\[6pt]
\textbf{pos}       & $\mat{p}_x^{(T)}$          &
$\displaystyle
  \mathcal L_{pos}
  =D_{\mathrm{KL}}\!\bigl(\mat{p}_x^{(T)}\;\|\;|\bm{\psi}_\Theta^{(T)}|^2\bigr)$ \\[6pt]
\textbf{pos\,\&\;mom} & $\mat{p}_x^{(T)},\,\mat{p}_{\xi}^{(T)}$ &
$\displaystyle
  \mathcal L_{pos\&mom}
  = D_{\mathrm{KL}}\!\bigl(\mat{p}_x^{(T)}\;\|\;|\bm{\psi}_\Theta^{(T)}|^2\bigr)
  + D_{\mathrm{KL}}\!\bigl(\mat{p}_{\xi}^{(T)}\;\|\;|\widehat{\bm{\psi}}_\Theta^{(T)}|^2\bigr)$\\
\bottomrule
\end{tabular}
\end{center}

%% file: appendix_proof.tex
\subsection{Notation and preliminaries}\label{app:proof_notation}
Throughout, $\mathbb{T}^d:=[-\pi,\pi]^{d}$ denotes the flat $d$–torus and  
$\mathbb{Z}^d$ the lattice of Fourier indices.  
For $\bm{\xi}\in\mathbb{Z}^{d}$ let $e_{\bm{\xi}}(\mat{x})\!:=\!e^{i\bm{\xi}\cdot \mat{x}}$.
The Fourier coefficient of a square integrable function $\mat{f}$ is  
\[
  \widehat{\mat{f}}(\bm{\xi})\;:=\;(2\pi)^{-d}\!\int_{\mathbb{T}^{d}} \mat{f}(\mat{x})\,e^{-i\bm{\xi}\cdot \mat{x}}\,\mathrm{d\mat{x}} .
\]
In a Sobolev space $H^s$ with an order of smoothness $s\in\mathbb{R}$, the Sobolev norm of function $\mat f$ is
\[
  \|\mat{f}\|_{H^{s}}^{2}\;:= \frac{(2\pi)^d}{2} \sum_{\bm{\xi}\in\mathbb{Z}^{d}}(1+|\bm{\xi}|^{2s})\,|\widehat{\mat{f}}(\bm{\xi})|^{2}.
\]

\paragraph{Asymptotics.}
Write $A\lesssim B$ if $A\le C\,B$ for a constant $C$ depending only on fixed
parameters (dimension, regularity exponents, etc.).

\paragraph{Vector Notation}
Fix a spatial dimension $d\ge 1$ and an index $j\in\{1,\dots,d\}$.
For $\bm{\xi}=(\xi_1,\dots,\xi_d)\in\mathbb{Z}^d$ we write
\[
  \bm{\xi}_{-j}:=(\xi_1,\dots,\xi_{j-1},\xi_{j+1},\dots,\xi_d)\in\mathbb{Z}^{d-1}
\]
for the vector obtained by \emph{removing} the $j$-th coordinate of $\bm{\xi}$.  
Conversely, for $\bm\alpha=(\alpha_1,\dots,\alpha_{d-1})\in\mathbb{Z}^{d-1}$ and $n\in\mathbb{Z}$ we define the \emph{insertion} map
\[
  (\bm\alpha,n)_j := (\alpha_1,\dots,\alpha_{j-1},\,n,\,\alpha_j,\dots,\alpha_{d-1})\in\mathbb{Z}^d .
\]
We use $|\cdot|_\infty$ for the max–norm on $\mathbb{Z}^d$, i.e.\ $|\bm{\xi}|_\infty=\max_{1\le i\le d}|\xi_i|$.
When we write $\widehat{\mat u}(\bm\alpha,r)$, this is shorthand for the $d$–dimensional coefficient
$\widehat{\mat u}\big((\bm\alpha,r)_j\big)$.

\medskip

\subsection{Proof of Lemma~\ref{lem:position-lower}}
\label{app:lem1proof}
\paragraph{Restatement of Lemma~\ref{lem:position-lower} (with explicit notation).}
Let $\mat u\in H^{s}(\mathbb{T}^{d})$ with $s>0$, and assume that its Fourier coefficients are compactly supported:
\[
  \widehat{\mat u}(\bm{\xi})=\mat 0 \quad\text{for all } |\bm{\xi}|_\infty>N_{0},
  \qquad\text{and}\qquad \widehat{\mat u}\not\equiv \mat 0 .
\]
Fix $j\in\{1,\dots,d\}$ and set $\mat v(\mat x):=x_j\,\mat u(\mat x)$.
Then there exist
\[
  \bm{\alpha}\in\mathbb{Z}^{d-1},\qquad
  m\in\{1,2,\dots,2N_{0}+1\},\qquad
  c>0,\quad R\in\mathbb{N},
\]
and an infinite set of the fiber at $j^{\text{th}}$ coordinate,
\[
  \Xi_{\bm{\alpha},R}\subset\bigl\{\bm{\xi}\in\mathbb{Z}^{d}:\ \bm{\xi}_{-j}=\bm{\alpha},\ |\xi_{j}|\ge R\bigr\}
\]
such that
\begin{equation}\label{eq:lem1_ineq}
  \bigl|\widehat{\mat v}(\bm{\xi})\bigr|
  \;\ge\; \frac{c}{\bigl(1+|\xi_{j}|\bigr)^{\,m}}
  \qquad \forall\,\bm{\xi}\in\Xi_{\bm{\alpha},R} .
\end{equation}
In particular, if for some $\bm{\alpha}\in\mathbb{Z}^{d-1}$ one has
\[
  \sum_{r=-N_{0}}^{N_{0}} \widehat{\mat u}\bigl((\bm{\alpha},r)_j\bigr)\neq \mat 0
  \quad\text{(equivalently, }\sum_{r=-N_{0}}^{N_{0}}\widehat{\mat u}(\bm{\alpha},r)\neq\mat 0\text{ in the shorthand above),}
\]
then the bound \eqref{eq:lem1_ineq} holds with the sharper exponent $m=1$.

\begin{proof}
We work with square-integrable function $\mat v\in L^{2}(\mathbb{T}^{d})$ where $\mathbb{T}^{d}=[0,2\pi]^{d}$. 
\medskip

\noindent\textbf{Step 1 — The periodic ``coordinate'' and its Fourier coefficients.}
The function $\mat x\mapsto x_{j}$ is not periodic.  Introduce the zero-mean, periodic 1D sawtooth
\[
\psi_{j}(\mat x)\;:=\;x_{j}-\pi,\qquad \mat x\in[0,2\pi]^{d},
\]
extended periodically to $\mathbb{T}^{d}$.  A direct computation (factorization of the integral and one-dimensional
integration by parts) shows that its Fourier coefficients are supported on the $j^{\text{th}}$-coordinate:
for $\bm{\xi}\in\mathbb{Z}^{d}$,
\begin{equation}\label{eq:psi-hat}
\widehat{\psi_{j}}(\bm{\xi})
=\begin{cases}
-\dfrac{1}{i\,\xi_{j}}, & \text{if } \bm{\xi}_{-j}=\bm{0}\ \text{and } \xi_{j}\neq 0,\\[0.75em]
0,& \text{if } \bm{\xi}=\bm{0}\ \text{or } \bm{\xi}_{-j}\neq \bm{0}.
\end{cases}
\end{equation}
Moreover $x_{j}=\psi_{j}+\pi$, hence
\begin{equation}\label{eq:v-split}
\widehat{\mat v}(\bm{\xi})\;=\;\ten{F}[\psi_{j}\mat u](\bm{\xi})+\pi\,\widehat{\mat u}(\bm{\xi}).
\end{equation}

\medskip
\noindent\textbf{Step 2 — Exact coefficient formula outside the support of $\widehat{\mat u}$.}
Since $\widehat{\mat u}(k)=\mat 0$ for $|\bm{\xi}|_{\infty}>N_{0}$, the second term in~\eqref{eq:v-split}
vanishes whenever $|\bm{\xi}|_{\infty}>N_{0}$. Using~\eqref{eq:psi-hat} and the convolution theorem, we obtain for any
$\bm{\xi}\in\mathbb{Z}^{d}$ with $|\bm{\xi}|_{\infty}>N_{0}$:
\begin{equation}\label{eq:tail-line}
\widehat{\mat v}(\bm{\xi})
=\sum_{\bm{\ell}\in\mathbb{Z}^{d}}\widehat{\psi_{j}}(\bm{\xi}-\bm{\ell})\,\widehat{\mat u}(\bm{\ell})
= -\frac{1}{i}\sum_{\substack{\bm{\ell}_{-j}=\bm{\xi}_{-j}\\ |\ell_{j}|\le N_{0}}}
\frac{\widehat{\mat u}(\bm{\ell})}{\xi_{j}-\ell_{j}}.
\end{equation}
Thus, along any fixed transverse index $\bm{\alpha}:=\bm{\xi}_{-j}\in\mathbb{Z}^{d-1}$, the tail $\widehat{\mat v}(\bm{\alpha},n)$ for large
$|n|$ is a finite sum of simple fractions in the single variable $n$.

\medskip
\noindent\textbf{Step 3 — Moments on a nontrivial fiber and the first non-vanishing moment.}
Because $\widehat{\mat u}\not\equiv\mat 0$, there exists at least one $\bm{\alpha}\in\mathbb{Z}^{d-1}$ for which the fiber
\[
\mathcal{F}_{\bm\alpha}:=\{r\in\mathbb{Z}:\;\widehat{\mat u}(\bm\alpha,r)\neq\mat 0\}
\]
is nonempty.
Define the (finite) coefficients $\mat{c}_{r}:=\widehat{\mat u}(\bm\alpha,r)$ for $r\in[-N_{0},N_{0}]$, and their moments
\[
\bm\mu_{p}:=\sum_{r=-N_{0}}^{N_{0}} r^{p}\mat{c}_{r}\qquad (p\ge 0).
\]
Let $m\in\{1,\dots,2N_{0}+1\}$ be the smallest index for which $\mu_{m-1}\neq\mat 0$.
Such an $m$ exists since not all $\mat{c}_{r}$ vanish.

\medskip
\noindent\textbf{Step 4 — Asymptotics along a line and a polynomial lower bound.}
For $n\in\mathbb{Z}$ with $|n|>N_{0}$, formula~\eqref{eq:tail-line} along the line $\bm{\xi}_{-j}=\bm\alpha$ reads
\[
\widehat{\mat v}(\bm\alpha,n)=-\frac{1}{i}\sum_{r=-N_{0}}^{N_{0}}\frac{\mat{c}_{r}}{n-r}.
\]
Expanding $\frac{1}{n-r}=\frac{1}{n}\sum_{q\ge 0}\bigl(\frac{r}{n}\bigr)^{q}$ for $|n|>2N_{0}$ and collecting
terms yields the asymptotic expansion
\[
\widehat{\mat v}(\bm\alpha,n)
=-\frac{1}{i}\left(\frac{\bm\mu_{0}}{n}+\frac{\bm\mu_{1}}{n^{2}}+\cdots+\frac{\bm\mu_{m-1}}{n^{m}}+O\!\left(\frac{1}{|n|^{m+1}}\right)\right),
\qquad |n|\to\infty.
\]
By the choice of $m$, the first nonzero term is $\bm\mu_{m-1}/n^{m}$. Consequently, there exist
$R\in\mathbb{N}$ and $c>0$ such that
\[
\bigl|\widehat{\mat v}(\bm\alpha,n)\bigr|
\;\ge\;\frac{c}{|n|^{m}}\qquad\text{for all }|n|\ge R,\; n\in\mathbb{Z}.
\]

\medskip
\noindent\textbf{Step 5 — Conclusion and the special case $m=1$.}
Then~\eqref{eq:lem1_ineq} holds for all $\bm{\xi}\in\Xi_{\bm\alpha, R}$ with the exponent $m$ determined in Step~3, by the definition of $\Xi_{\bm\alpha, R}$ and that $|n|\ge R$.
If $\bm\mu_{0}=\sum_{r}\mat{c}_{r}\neq\mat 0$ for the chosen fiber (equivalently,
$\sum_{r=-N_{0}}^{N_{0}}\widehat{\mat u}(\bm\alpha,r)\neq\mat 0$), then $m=1$ and we obtain the sharper
$|\widehat{\mat v}(\bm{\xi})|\gtrsim(1+|\xi_{j}|)^{-1}$ along the line $\bm{\xi}_{-j}=\bm\alpha$.
\end{proof}

\begin{remark}
The explicit one-dimensional tail~\eqref{eq:tail-line} shows that multiplying a band-limited field
by the coordinate $x_{j}$ produces a \emph{polynomial} Fourier tail decay along lines parallel to the $j^{\text{th}}$ coordinate,
with rate $(1+|\xi_{j}|)^{-m}$ where $m$ is the first non-vanishing moment of the finitely many coefficients on
the relevant fiber. In particular, when $m=1$, the decay is exactly $(1+|\xi_{j}|)^{-1}$.
Such algebraic tails are consistent with the pseudo-spectral projection error estimate quoted in ~\citet[Thm.\;40]{uatfno}. 
\end{remark}

\subsection{Proof of Theorem~\ref{thm:fno-fail}}
\label{app:thm1proof}

\paragraph{Restatement of Theorem~\ref{thm:fno-fail}.}
Let $\bm{\alpha}=(\alpha_{1},\dots,\alpha_{d})\in\mathbb{N}^{d}$ with
total degree $M:=|\bm{\alpha}|\ge 1$ and define the position‑multiplier
\[
   \ten M(\mat x):=x_{1}^{\alpha_{1}}\,x_{2}^{\alpha_{2}}\cdots x_{d}^{\alpha_{d}}.
\]
For inputs band‑limited to radius $N_{0}$ and lying in
$H^{s}(\mathbb{T}^{d})$ with $s>\tfrac{d}{2}$, any Fourier Neural
Operator $\ten{G}^{\tiny{\text{FNO}}}_{\theta}$ that achieves
\(
   \|\ten M(\mat x)-\ten{G}^{\tiny{\text{FNO}}}_{\theta}\|_{H^{s}\!\to H^{s-\delta}}\le\varepsilon
\)
($0<\delta<1$)
must employ a spectral bandwidth (FNO width)
\(
  m\gtrsim\varepsilon^{-M/s}
\)
and a parameter count
\(
  |\theta|\;\ge\;\exp\!\bigl(c\,\varepsilon^{-Md/s}\bigr)
\)
for some $c>0$ depending only on $(d,s,\delta,N_{0})$.

\begin{proof}
    
Set \(s' := s-\delta\) with \(0<\delta<1\).

\medskip
\noindent\textbf{Step 1 – Algebraic tail produced by \(\ten M(\mat x)\).}  
Applying Lemma~\ref{lem:position-lower} once per factor of \(x_{j}\) shows that
for some constant \(C_{0}>0\) and an infinite set
\(\Xi_\infty\subset\mathbb{Z}^{d}\),
\begin{equation}\label{eq:tail-poly}
  |\ten{F}[\ten M(\mat x)\mat{u}](\bm{\xi})|
  \;\ge\;
  \frac{C_{0}}{(1+|\bm{\xi}|)^{M+1}}
  \qquad\forall\,\bm{\xi}\in\Xi_\infty.
\end{equation}

\medskip
\noindent\textbf{Step 2 – Pseudo‑spectral projection lower bound.}  
For any \(\mat f\in H^{s}\), pseudo-spectral projection error estimate gives~\citep{spectraltxt}
\[
  \|(\ten{I}-\bm{\Pi}_{\tiny{\text{FNO}}})\mat f\|_{H^{s'}}
  \;\ge\;
  C_{1}\!
  \Bigl[\!\!
     \sum_{|\bm{\xi}|_{\infty}>N}
     (1+|\bm{\xi}|^{2s'})\,|\widehat{\mat f}(\bm{\xi})|^{2}
  \Bigr]^{1/2},
\]
where $\ten{I}$ is an identity operator. Insert \(\mat f=\ten M(\mat x)\mat u\) and the tail bound \eqref{eq:tail-poly}; summing over
\(\Xi_\infty\cap\{|\bm{\xi}|_{\infty}>m\}\) yields
\[
  \|(\ten{I}-\bm{\Pi}_{\tiny{\text{FNO}}})\ten M(\mat x)\mat u\|_{H^{s'}}
  \;\gtrsim\;
  m^{-(M-\delta)}.
\]
Imposing this residual \(\le \tfrac12\varepsilon\) forces
\begin{equation}\label{eq:N-bound}
  m\;\ge\;C_{2}\,\varepsilon^{-M/s},
  \qquad  C_{2}=C_{2}(d,s,\delta,M,N_{0})>0 .
\end{equation}

\medskip
\noindent\textbf{Step 3 – Canonical neural scaling in the latent map.}  
An FNO with spectral radius (width) \(m\) manipulates a latent vector of
dimension \((2m+1)^{d}\sim m^{d}\).
Approximating a generic Lipschitz map
\(\mat{G}: \mathbb{C}^{m^{d}}\!\to\mathbb{C}^{m^{d}}\)
to accuracy \(\varepsilon/2\) with a fully connected network requires~\citep{reluscaling,tanhscaling}
\emph{at least}
\[
  \text{parameters}
  \;\gtrsim\;
  \varepsilon^{-m^{d}}.
\]
\citep[Remark 22]{uatfno}

\medskip
\noindent\textbf{Step 4 – Substitute the bandwidth lower bound.}  
Using \eqref{eq:N-bound},
\(
  m^{d}\sim \varepsilon^{-M d/s}.
\)
Hence the latent network must have at least
\[
  |\theta|
  \;\gtrsim\;
  \varepsilon^{-\varepsilon^{-M d/s}},
\]
a \emph{super‑exponential} curse of dimensionality in the target accuracy
\( \varepsilon \).

\end{proof}

\subsection{Proof of Theorem~\ref{thm:kano-general}}
\label{app:thm2proof}

\paragraph{Restatement of Theorem~\ref{thm:kano-general}.}
Let \(s\ge s'\ge 0\) and \(s_{p}>d/2\).
Consider any finite composition
\[
   \ten{G} \;=\; (\varsigma_{\ell}\!\circ \ten{L}_{\ell})\circ \cdots \circ (\varsigma_{1}\!\circ \ten{L}_{1}),
   \qquad
   \ten{L}_{i}= \mat{Op}_m(\boldsymbol{p}_{i}),
\]
where each Kohn–Nirenberg symbol \(\boldsymbol{p}_{i}\) belongs to
\(W^{s_{p},2}\bigl(\mathbb{T}^{d}_{\mat x}\!\times\!\mathbb{T}^{d}_{\bm{\xi}}\bigr)\)\footnote{Standard square integrable periodic Sobolev space on the product torus \(\mathbb{T}^{2d}\).},
and each \(\varsigma_i\) is a point‑wise nonlinearity of uniformly bounded degree
(so the number of such nonlinearities is \(O(\ell)\) and independent of \(\varepsilon\)).
For every \(\varepsilon>0\) there exists a \emph{single‑layer} KANO
\(\ten{G}^{\tiny{\text{KANO}}}_{\theta}\) such that
\[
   \|\ten{G}-\ten{G}^{\tiny{\text{KANO}}}_{\theta}\|_{H^{s}\!\to H^{s'}}\le\varepsilon,
   \qquad
   |\theta|\;=\;
   O\!\bigl(\varepsilon^{-d/(2s_{p})}\bigr).
\]

\begin{proof}

\noindent\textbf{Step 1 – Kohn–Nirenberg quantization error.}
By the quadrature bound from~\citet{dsc}, for each \(\ten L_{i}\) we have
\[
  \bigl\|\ten L_{i} - \mat{Op}_{m}(\boldsymbol{p}_{i})\bigr\|_{H^{s}\!\to H^{s'}}
  \;\le\; C'\,B\,m^{-(s-s')},
\]
where \( \mat{Op}_{m}(\boldsymbol{p}_{i}) \) keeps only frequencies
\(|\bm{\xi}|_{\infty}\le m\).
Since each \(\varsigma_i\) is a bounded‑degree pointwise map, its Nemytskii operator is Lipschitz on bounded sets; write \(L_i:=\mathrm{Lip}(\varsigma_i)\)\footnote{Lipschitz constant of the point-wise nonlinearity $\varsigma_i$ on the relevant value range, i.e. $L_i:=\sup_{a\neq b}\frac{|\varsigma_i(a)-\varsigma_i(b)|}{|a-b|}$ with $a,b$ restricted to the compact interval attained by the $i$-th preactivations.} on the relevant range and set \(L_{\max}:=\max_i L_i\).
Let
\(M:=\max_j\!\bigl\{\|\ten L_j\|_{H^{s'}\!\to H^{s'}},\,\|\mat{Op}_{m}(\boldsymbol{p}_{j})\|_{H^{s'}\!\to H^{s'}}\bigr\}\).
Because \(s_p>\tfrac d2\) implies \(\|\boldsymbol{p}_{j}\|_{L^{\infty}}\!\lesssim\!\|\boldsymbol{p}_{j}\|_{W^{s_p,2}}\),
both \(\ten L_j\) and \(\mat{Op}_{m}(\boldsymbol{p}_{j})\) are bounded on \(H^{s'}\) with a bound independent of \(m\).
A telescoping estimate for the interleaved composition then yields
\[
  \bigl\|\ten G-\bm{\Pi}_{\tiny{\text{KANO}}}(\ten G)\bigr\|_{H^{s}\!\to H^{s'}}
  \;\le\; C_\ast\,C'\,B\,m^{-(s-s')},\qquad
  C_\ast\;\le\;\ell\,(L_{\max}M)^{\ell-1}L_{\max}.
\]

Pick \( m := \bigl(2\,C_\ast\,C'B/\varepsilon\bigr)^{1/(s-s')}\).
Then the full composition \(\ten G\) deviates from its projected counterpart
\[
  \bm{\Pi}_{\tiny{\text{KANO}}}(\ten G)
  := (\varsigma_{\ell}\!\circ \mat{Op}_{m}(\boldsymbol{p}_{\ell}))\circ\cdots\circ
     (\varsigma_{1}\!\circ \mat{Op}_{m}(\boldsymbol{p}_{1}))
\]
by at most \( \varepsilon/2 \) in \(H^{s}\!\to H^{s'}\) operator norm.

\medskip
\noindent\textbf{Step 2 – KAN approximation of both symbols and pointwise nonlinearities.}
By the width‑fixed KAN scaling law from~\citet{expressivitykan}, for any \(\eta>0\) there exists
a Kolmogorov–Arnold Network \(\bm{\Phi}_{i,\eta}\) with
\(|\bm{\Phi}_{i,\eta}|=O(\eta^{-d/(2s_{p})})\) such that
\(\|\boldsymbol{p}_{i}-\bm{\Phi}_{i,\eta}\|_{L^{\infty}}\le\eta\) on
\(\mathbb{T}^{d}_{\mat x}\!\times\!\mathbb{T}^{d}_{\bm{\xi}}\).
Likewise, since each \(\varsigma_i\) is a fixed‑degree pointwise map, its action over the compact
value range visited by the projected flow can be uniformly approximated by a width‑fixed KAN
\(\bm{\Psi}_{i,\eta}\) with size
\(|\bm{\Psi}_{i,\eta}| = O\!\bigl(\eta^{-1/(2s_{p})}\bigr)\) and
\(\|\varsigma_i-\bm{\Psi}_{i,\eta}\|_{L^{\infty}}\le \eta\).
Choose
\[
   \eta_{\mathrm{sym}} := \frac{\varepsilon}{4\ell}, \qquad
   \eta_{\mathrm{nl}}  := \frac{\varepsilon}{4\ell}.
\]
Define the single‑layer KANO
\[
  \ten{G}^{\tiny{\text{KANO}}}_{\theta}
    := (\bm{\Psi}_{\ell,\eta_{\mathrm{nl}}}\!\circ \mat{Op}_{m}(\bm{\Phi}_{\ell,\eta_{\mathrm{sym}}}))\circ
       \cdots\circ
       (\bm{\Psi}_{1,\eta_{\mathrm{nl}}}\!\circ \mat{Op}_{m}(\bm{\Phi}_{1,\eta_{\mathrm{sym}}})),
\]
where the same spectral radius \(m\) from Step~1 is used in every \(\mat{Op}_{m}(\,\cdot\,)\).

\medskip
\noindent\textbf{Step 3 – Error accumulation beyond projection.}
Using linearity of the symbol‑to‑operator map and stability of Nemytskii (pointwise) maps under
uniform approximation, the post‑projection error splits into a sum of the symbol parts and the
nonlinearity parts:
\begin{align*}
  \bigl\|\bm{\Pi}_{\tiny{\text{KANO}}}(\ten G)-\ten{G}^{\tiny{\text{KANO}}}_{\theta}\bigr\|_{H^{s}\!\to H^{s'}}
  &\le
  \sum_{i=1}^{\ell}
    \Bigl\|\mat{Op}_{m}(\boldsymbol{p}_{i})-\mat{Op}_{m}(\bm{\Phi}_{i,\eta_{\mathrm{sym}}})\Bigr\|_{H^{s}\!\to H^{s'}}
  \;+\;
  \sum_{i=1}^{\ell}
    \|\varsigma_i-\bm{\Psi}_{i,\eta_{\mathrm{nl}}}\|_{L^{\infty}}
  \\
  &\le \ell\,\eta_{\mathrm{sym}} + \ell\,\eta_{\mathrm{nl}}
   \;=\; \varepsilon/2.
\end{align*}
Combining with Step~1 yields
\(
  \|\ten{G}-\ten{G}^{\tiny{\text{KANO}}}_{\theta}\|_{H^{s}\!\to H^{s'}} \le \varepsilon.
\)

\medskip
\noindent\textbf{Step 4 – Parameter complexity.}
Summing the sizes of all KAN blocks gives
\[
  |\theta|
  \;=\;
  \sum_{i=1}^{\ell} O\!\bigl(\eta_{\mathrm{sym}}^{-d/(2s_{p})}\bigr)
  \;+\;
  \sum_{i=1}^{\ell} O\!\bigl(\eta_{\mathrm{nl}}^{-1/(2s_{p})}\bigr)
  \;+\; O(1)
  \;=\;
  O\!\bigl((\varepsilon/\ell)^{-d/(2s_{p})}\bigr) + O\!\bigl((\varepsilon/\ell)^{-1/(2s_{p})}\bigr).
\]
Since \(\ell\) and the number/degree of the \(\varsigma_i\) are fixed (do not scale with \(\varepsilon\)),
the dominating term is \(O(\varepsilon^{-d/(2s_{p})})\), establishing the claimed complexity.
\end{proof}

%% file: appendix_high_order.tex
In Section~\ref{sec:fno_example} we performed a linear analysis of a single FNO layer via its Jacobian to illustrate the pure-spectral bottleneck. We showed that, in the first-order approximation, all spectral off–diagonals arise from the spectrum of the input–dependent gate $\sigma'\!\big(\mat{z}(\mat{u},\cdot)\big)$~(\ref{eq:fno-jacobian-fourier}). Hence, although a single FNO layer is capable of generating spectral off-diagonals, they are tied to the input distribution and leads to the structural fragility in out-of-distribution performance (generalization on the unseen input distribution). Experimental results in Section~\ref{sec:experiments} are aligned with our concerns. 

In this Appendix, we show that the same phenomenon persists at \emph{every} order of the Fr\'echet expansion of a single FNO layer, and composing deep layers does not remove the input distribution dependence of the spectral off-diagonals generated by the model as well.

Throughout, we work on the flat torus $\mathbb{T}^d = [-\pi,\pi]^d$ and Sobolev spaces $H^s$ as in Appendix~\ref{app:proofs}.

\subsection{Higher–order spectral off–diagonals of FNO}\label{app:fno-higher-order}

Recall the FNO layer $\ten{L}_{\text{\tiny FNO}}$~(\ref{eq:fno_layer}):
\[
    \ten{L}_{\text{\tiny{FNO}}}(\mat{u})(\mat{x})\;=\;\sigma\!\left(
\ten{F}_m^{-1}\!\Big(\mat{R}_{\theta_1}(\bm{\xi})\cdot\;\ten{F}_m(\mat{u})(\bm{\xi})\Big)(\mat{x})\;+\;
\mat{W}_{\theta_2} \cdot \mat{u}(\mat{x})
\right).
\]
For the analysis it is convenient to collect the linear terms into a linear operator
\begin{equation}\label{eq:app-A-def}
    \ten{A} \;:=\; \ten{F}_m^{-1} \circ \mat{R}_\theta(\bm{\xi}) \circ \ten{F}_m \;+\; \mat{W}_\theta,
\end{equation}
so that
\begin{equation}\label{eq:app-F-def}
    \ten{L}_{\text{\tiny FNO}}(\mat{u})(\mat{x})
    \;=\;
    \sigma\!\big(\mat{z}(\mat{u},\mat{x})\big),
    \qquad
    \mat{z}(\mat u,\mat x) := (\ten A \mat u)(\mat x),
\end{equation}
where $\sigma$ is the non-linear activation acting point-wise.

We first compute the $k$-th Fr\'echet derivative~\citep{frechet} of $\ten{L}_{\text{\tiny FNO}}(\mat{u})(\mat{x})$.

\begin{proposition}[\textbf{Structure of higher–order derivatives of a single Fourier layer}]\label{prop:AppC5-derivatives}
Let $\mat u \in H^s$ and $k$ be the order of the Fre\'echet derivative. Also let $\mat h_1,\dots,\mat h_k \in H^s$ be the arbitrary direction functions of each order. Then for every integer $k\ge 1$, the Fre\'echet derivative of $\ten{L}_{\text{\tiny FNO}}(\mat{u})(\mat{x})$ is:
\begin{equation}\label{eq:app-F-kth-derivative}
    \ten{D}^k \ten{L}_{\text{\tiny FNO}}(\mat{u})[\mat h_1,\dots,\mat h_k](\mat x)
    \;:=\;
    \sigma^{(k)}\!\big(\mat z(\mat u,\mat x)\big)\,
    \prod_{j=1}^k (\ten A \mat h_j)(\mat x),
\end{equation}
where $\sigma^{(k)}$ is the usual scalar $k$-th derivative. More explicitly, for each channel $q$,
\begin{equation}\label{eq:app-F-kth-derivative-channel}
    \big[\ten{D}^k \ten{L}_{\text{\tiny FNO}}(\mat{u})[\mat h_1,\dots,\mat h_k]\big]_q(\mat x)
    \;:=\;
    \sigma^{(k)}\!\big(\mat z_q(\mat u,\mat x)\big)\,
    \prod_{j=1}^k [\ten A \mat h_j]_q(\mat x).
\end{equation}
\end{proposition}

\begin{proof}
Since $\ten A$~(\ref{eq:app-A-def}) is linear, $\ten D \mat z(\mat u) \equiv \ten A$ and $\ten D^r \mat z(\mat u) \equiv 0$ for all $r\ge 2$. By the chain rule,
\[
    \ten D\ten{L}_{\text{\tiny FNO}}(\mat{u})[\mat h_i](\mat x)
    \;=\;
    \sigma'\!\big(\mat z(\mat u,\mat x)\big)\,(\ten A \mat h_i)(\mat x),
\]
where $\sigma'$ acts point-wise on each channel. For higher order derivatives, we apply the Fa\`a~di~Bruno formula~\citep{faadibruno}. Because all higher derivatives of $\mat z$ vanish, every term involving $\ten D^r \mat z(\mat u)$ with $r\ge 2$ drops out, leaving
\begin{equation}
    \ten D^k \ten{L}_{\text{\tiny FNO}}(\mat{u})[\mat h_1,\dots,\mat h_k]\;=\;\sigma^{(k)}(\mat z(\mat u))\big[\ten A \mat h_1,\dots,\ten A \mat h_k\big].
\end{equation}
Since $\sigma$ acts point-wise, this reduces to Eq.~(\ref{eq:app-F-kth-derivative}), and to Eq.~(\eqref{eq:app-F-kth-derivative-channel}) on each channel.
\end{proof}

Thus, as apparent in Eq.~(\ref{eq:app-F-kth-derivative-channel}), \emph{every} Fr\'echet derivative is a point-wise product of 
\begin{itemize}
    \item an input-dependent gate $\sigma^{(k)}\!\big(\mat z(\mat u,\mat x)\big)$ depending on the current input $\mat u$, and
    \item the product of $k$ fixed linear responses $(\ten A\mat h_i)(\mat x)$ incapable of generating spectral off-diagonals.
\end{itemize}
Therefore, the structure of the high-order derivatives of a single layer FNO is just as tied to the input $\mat u$ distribution as the first-order Jacobian shown in Section~\ref{sec:fno_example}.

\subsection{Spectral Representation of High-Order Derivatives and Multi-Layer FNO}

Fix $k\ge 1$ and define
\begin{equation}
    \mat s_k(\mat u,\mat x):=\sigma^{(k)}\big(\mat z(\mat u,\mat x)\big),
    \qquad
    \mat b_i(\mat x) := (\ten A \mat h_i)(\mat x),\quad i=1,\dots,k.
\end{equation}

Then Eq.~(\ref{eq:app-F-kth-derivative}) reads
\begin{equation}\label{eq:app-F-kth-pointwise}
    \ten D^k \ten{L}_{\text{\tiny FNO}}(\mat{u})[\mat h_1,\dots,\mat h_k](\mat x)
    \;=\;
    \mat s_k(\mat u,\mat x)\,\cdot\prod_{i=1}^k \mat b_i(\mat x).
\end{equation}
The Fourier transform of a product of $(k{+}1)$ functions is a $(k{+}1)$-fold convolution, so for each output frequency $\bm{\xi}\in\mathbb Z^d$ we have
\begin{equation}\label{eq:app-F-kth-fourier-convolution}
    \widehat{\ten{D}^k \ten{L}_{\text{\tiny FNO}}(\mat{u})[\mat h_1,\dots,\mat h_k]}(\bm{\xi})
    \;=\;
    \big(\widehat{\mat s_k(\mat u)} * \widehat{\mat b_1} * \cdots * \widehat{\mat b_k}\big)(\bm{\xi}).
\end{equation}
Spectrally, the linear operator $\ten A$~(\ref{eq:app-A-def}) acts diagonally:
\begin{equation}\label{eq:app-A-fourier}
    \widehat{\ten A \mat h_i}(\bm{\xi})
    \;=\;
    \mat A(\bm{\xi})\,\widehat{\mat h_i}(\bm{\xi}),
\end{equation}
where $\mat A(\bm{\xi})=\mat R(\bm{\xi}) + \mat W$ as in Eq.~(\ref{eq:fno-jacobian-fourier}). Expanding the convolution in Eq.~(\ref{eq:app-F-kth-fourier-convolution}) yields
\begin{equation}\label{eq:app-F-kth-fourier-expanded}
    \widehat{\ten D^k \ten{L}_{\text{\tiny FNO}}(\mat{u})[\mat h_1,\dots,\mat h_k]}(\bm{\xi})
    \;=\;
    \sum_{\bm{\xi}_1,\dots,\bm{\xi}_k \in \mathbb Z^d}
    \underbrace{\widehat{\mat s_k(\mat u)}\big(\bm{\xi} - \bm{\xi}_1 - \cdots - \bm{\xi}_k\big)}_{\text{depends on input}\mat u}
    \;\cdot\;\prod_{i=1}^k \mat A(\bm{\xi}_i)\,\widehat{\mat h_i}(\bm{\xi}_i),
\end{equation}
where $\{\bm{\xi}_i\}$ are the frequency variable for each direction $\{\mat h_i\}$. 
Eq.~(\ref{eq:app-F-kth-fourier-expanded} shows that the $k$-th derivative could be read as a multi-linear operator whose \emph{spectral kernel}
\begin{equation}\label{eq:app-F-kth-kernel}
    \mat K_k(\mat u;\bm{\xi};\bm{\xi}_1,\dots,\bm{\xi}_k)
    \;:=\;
    \widehat{\mat s_k(\mat u)}\big(\bm{\xi} - \bm{\xi}_1 - \cdots - \bm{\xi}_k\big)\,
    \cdot\prod_{i=1}^k \mat A(\bm{\xi}_i)
\end{equation}
connects $(\bm{\xi}_1,\dots,\bm{\xi}_k)$ to the output function frequency $\bm{\xi}$.

Two structural facts follow immediately from Eq.~(\ref{eq:app-F-kth-kernel}):
\begin{enumerate}[leftmargin=*]
    \item Input function $\mat u$ decides the value $\widehat{\mat s_k(\mat u)}=\ten F\big[\sigma^{(k)}\big(\mat z(\mat u,\cdot)\big)\big],$
    \item The spectral multipliers $\mat A(\bm{\xi}_i)$ depend only on the learned weights and remain diagonal in Fourier: they are independent of input $\mat u$ but does not contribute to spectral off-diagonals.
\end{enumerate}

Consequently, for any $k\ge 1$:
\begin{quote}
\emph{Every non-diagonal spectral coupling is induced by Fourier coefficients of non-linear activation derivatives $\sigma^{(k)}\!\big(\mat z(\mat u,\cdot)\big)$ evaluated on the current input $\mat u$.}
\end{quote}
Higher orders $k\ge 2$ introduce convolutions of higher derivatives $\sigma^{(k)}$, but they do not mitigate the input dependence argument of spectral off-diagonals provided in Section~\ref{sec:fno_example}.

\paragraph{Deep FNO networks.}
A full FNO is a composition of multiple FNO layers:
\begin{equation}
    \ten G^{\text{\tiny FNO}}_\theta
    \;=\; \ten L^{(\ell)}_{\text{\tiny FNO}} \circ \cdots \circ \ten L^{(1)}_{\text{\tiny FNO}}.
\end{equation}
 Applying the Fa\`a~di~Bruno formula~\citep{faadibruno} to this composition, every term in $\ten D^k \ten G^{\text{\tiny FNO}}_\theta(\mat u)$ becomes a product of:
\begin{itemize}
    \item spectral multipliers $\{\mat A_j(\bm{\xi})\}$ from the linear parts of the layers, and
    \item Fourier transforms of activation derivatives $\sigma^{(r)}\big(\mat z^{(j)}(\mat u,\cdot)\big)$ from intermediate layers $j$ and derivative orders $r\ge 1$.
\end{itemize}
Thus, in a \emph{deep} FNO, all spectral off-diagonals at any order $k$ still factor through Fourier transforms of activation derivatives evaluated on intermediate pre-activations governed by the input $\mat u$. Adding layers introduces more such gates but do not mitigate the input dependence on spectral off-diagonals introduced in Section~\ref{sec:fno_example}.

\subsection{The Meaning of Input Dependence of Spectral Off-Diagonals When Learning a Spectrally Dense Ground-Truth Operators}

Let $\ten T$ be a fixed linear operator such as the position multiplier $a(x)\mapsto x^2 a(x)$ in Section~\ref{sec:fno_example}. In Fourier basis, $\ten T$ is represented by a dense Toeplitz matrix such as $\mat T_n[x^2]$ as in Eq.~(\ref{eq:toeplitz-x2}), and its Fr\'echet derivatives are
\[
    \ten D \ten T(\mat u) \equiv \ten T \quad \text{for all } \mat u,
    \qquad
    \ten D^k \ten T(\mat u) \equiv 0 \quad \text{for all } k\ge 2,
\]
i.e., the spectral off-diagonals of $\ten T$ are completely \emph{independent} of the input $\mat u$.

Suppose we wish to learn $\ten T$ with robust out-of-distribution performance over a Sobolev ball
\[
    \mathcal{B}_B := \big\{\mat u \in H^s : \|\mat u\|_{H^s} \le B\big\}.
\]
A natural notion of generalization is a small error in operator norm for an unseen Sobolev ball $\mathcal B_B$:
\begin{equation}\label{eq:app-operator-error}
    \sup_{\mat u\in\mathcal{B}_B}
    \frac{\big\|\ten G^{\text{\tiny FNO}}_\theta(\mat u) - \ten T \mat u\big\|_{H^s}}{\|\mat u\|_{H^s}}
    \;\le\; \varepsilon.
\end{equation}
This requires that on $\mathcal{B}_B$
\begin{itemize}
    \item the Jacobian $\ten D \ten G^{\text{\tiny FNO}}_\theta(u)$ stays close to $\ten T$, and
    \item all higher-order derivatives $\ten D^k \ten G^{\text{\tiny FNO}}_\theta(\ten u)$, $k\ge 2$, remain uniformly small (negligible).
\end{itemize}
Combining Eq.~(\ref{eq:app-F-kth-kernel}) with the composition structure above reveals a paradox:
\begin{enumerate}
    \item If the gates $\sigma^{(k)}\big(\mat z^{(j)}(\mat u,\cdot)\big)$ and their Fourier transforms $\widehat{\mat s_{k}^{(j)}(\mat u)}$ vary significantly with $\mat u\in\mathcal{B}_B$, then both the Jacobian and all higher-order kernels $\mat K_k(\mat u;\bm{\xi},\bm{\xi}_1,\dots,\bm{\xi}_k)$ vary with the input. In that case the effective spectral off–diagonals of $\ten G^{\text{\tiny FNO}}_\theta$ cannot coincide with a single, input-independent Toeplitz kernel across all $\mat u\in\mathcal{B}_B$. Any dense off-diagonal pattern learned from the train distribution will be tied to the train subspace and becomes fragile under distribution shift, as observed in our experiments in Section~\ref{sec:experiments}.
    \item If, on the other hand, we try to make these kernels effectively independent of $\mat u$ on $\mathcal{B}_B$, then Eq.~(\ref{eq:app-F-kth-derivative}) and Eq.(\ref{eq:app-F-kth-kernel}) force all activation derivatives to be nearly constant (for $k=1$) or nearly zero (for $k\ge 2$) on the relevant pre-activation range. In this regime, the network is forced into an almost linear operating region where
    \begin{itemize}
        \item $\sigma'$ is approximately constant, so $\widehat{\mat s_1(u)}$ is concentrated near zero frequency and $\ten D \ten L_{\text{\tiny FNO}}(\mat u)$ becomes (block-)diagonal in spectral basis; and
        \item $\sigma^{(k)}\approx 0$ for $k\ge 2$, so higher-order terms vanish.
    \end{itemize}
    The resulting layer effectively reduces to a spectral multiplier $\mat A(\bm{\xi})$ and cannot represent a dense Toeplitz map such as $\mat T_n[x^2]$ whose off-diagonals are non-trivial and fixed independent of inputs.
\end{enumerate}

In other words, for spectrally dense, position-dependent operators, FNO faces a fundamental trade-off:
\begin{itemize}
    \item it can use strongly input-dependent gates to synthesize spectral off-diagonals, but then those off-diagonals are necessarily tied to the input distribution, or
    \item it can suppress the input dependence of the gates to emulate a fixed operator, but then the spectral kernel collapses towards a diagonal (or nearly diagonal) multiplier.
\end{itemize}
In neither case does a practical-size FNO realize a fixed, input-independent dense Toeplitz kernel with robust out-of-distribution generalization for spectrally dense operators, even though a sufficiently large FNO \emph{can} parametrize the off-diagonals of the in-sample mapping on the training distribution.

\paragraph{What we do (and do not) claim about FNO.}
The generalized analysis in this Appendix refines the statement of Section~\ref{sec:fno_example}:
\begin{itemize}
    \item We \emph{do not} claim that FNO cannot generate spectral off-diagonals. Eq.~(\ref{eq:fno-jacobian-fourier}) and Eq.~(\ref{eq:app-F-kth-kernel}) show that at first and higher order, off-diagonals appear whenever the activation derivatives have non-trivial Fourier coefficients.
    \item We \emph{do} claim that for spectrally dense operators, these off-diagonals are \emph{always tied to input-dependent gates}. As a result, a large FNO can fit the off-diagonals of the in-sample mapping on the training subspace, but it cannot efficiently learn a fixed dense off-diagonals with robust out-of-distribution generalization on unseen function spaces which is exactly what we observe in our experiments in Section~\ref{sec:experiments}.
\end{itemize}
Stacking more layers introduces more input-dependent gates but does not create an input-independent spectral mixing mechanism, so depth does not mitigate this bottleneck.

\subsection{Why KANO Does Not Suffer the Same Problem}

By contrast, KANO directly learns an input-independent pseudo-differential symbol $\mat p_\theta(\mat x,\bm{\xi})$ in the dual bases via Eq.~(\ref{eq:kano-layer}). KANO has a spectral kernel capable of mode mixing as it is governed by spatial basis $\mat x$ (which is convolution in spectral basis) as well as the frequency mode $\bm{\xi}$ via symbol $\mat p_\theta(\mat x,\bm{\xi})$, not only by the non-linear activation gates tied to the input function $\mat u$. Once $\mat p_\theta$ is learned, it is shared across all input functions in the ball $\mathcal{B}_B$, including even the unseen function subspace. This dual-domain, symbol-based parameterization allows KANO to learn the fixed off-diagonals of spectrally dense, position-dependent operators with robust out-of-distribution generalization, as confirmed by our experiments in Section~\ref{sec:experiments}.

%% file: appendix_computation_memory_complexity.tex
In this Appendix we compare the memory and computation complexity of KANO compared to FNO. We first quantify the per–layer costs of a single FNO layer~(\ref{eq:fno_layer}) and a single KANO layer~(\ref{eq:kano-layer}), and then combine
them with the parameter complexity results of Section~\ref{sec:thm1} and Section~\ref{sec:thm2} to argue that, on the target class of variable–coefficient PDE and position–dependent dynamics,
the higher per–layer cost of KANO is compensated by better model size scaling.

Throughout this appendix we work on the discrete torus $\mathbb T^d$ with a uniform spatial grid
$\mathcal Y = \{\mat y_1,\dots,\mat y_m\}$ and a truncated Fourier set $\Xi = \{\bm{\xi}_1,\dots,\bm{\xi}_m\}$ as in
Section~\ref{sec:fno_background} and Section~\ref{sec:kanoarch}.

\subsection{Per–layer FLOPs and activation memory}

We measure complexity in floating–point operations (FLOPs) and activation memory per forward pass for a single layer of each model. Backward passes in standard automatic differentiation are assumed to be within a constant factor of the forward cost and do not change the asymptotic conclusions.

\paragraph{FNO layer.}
Consider a single FNO layer~(\ref{eq:fno_layer})
\[
    \ten{L}_{\text{\tiny{FNO}}}(\mat{a})(\mat{x})\;=\;\sigma\!\left(
\ten{F}_m^{-1}\!\Big(\mat{R}_{\theta_1}(\bm{\xi})\cdot\;\ten{F}_m(\mat{a})(\bm{\xi})\Big)(\mat{x})\;+\;
\mat{W}_{\theta_2} \cdot \mat{a}(\mat{x})
\right)
\]
with $C_{\text{in}}$ input channels and $C_{\text{out}}$ output channels. On a $d$-dimensional grid $\mathcal Y$ with $m^d$ coordinates, a single forward application of $\ten L_{\text{\tiny FNO}}$ has the following costs:
\begin{itemize}
    \item \emph{FFT and inverse FFT:}
    $\ten F_m$ and $\ten F_m^{-1}$ are applied channel–wise and cost
    \[
        \text{FFT cost} \;\sim\; \mathcal O\!\big((C_{\text{in}} + C_{\text{out}})\, m^d \log m^d\big).
    \]
    \item \emph{Spectral multiplier:} for each retained mode $\bm{\xi}\in\Xi$,
    $\mat R_\theta(\bm{\xi}) \in \mathbb C^{C_{\text{out}}\times C_{\text{in}}}$ is a dense matrix; multiplying by $\hat{\mat u}(\bm{\xi})\in\mathbb C^{C_{\text{in}}}$ costs $\mathcal O(C_{\text{in}} C_{\text{out}})$ per mode, hence
    \[
        \text{spectral block} \;\sim\; \mathcal O\!\big(m^d\,C_{\text{in}} C_{\text{out}}\big).
    \]
    \item \emph{Point-wise linear map:} $\mat W_\theta$ is applied at each spatial point $\mat y\in \mathcal Y$ as a
    dense matrix in channel space, costing
    \[
        \text{spatial linear map} \;\sim\; \mathcal O\!\big(m^d\,C_{\text{in}} C_{\text{out}}\big).
    \]
    \item \emph{Nonlinearity:} the point-wise nonlinearity $\sigma$ is $\mathcal O(m^d\,C_{\text{out}})$.
\end{itemize}
Collecting terms, for fixed channel counts we obtain the per–layer forward cost:
\begin{equation}\label{eq:fno-flops}
    \text{FLOPs}\big(\ten L_{\text{\tiny FNO}}\big)
    \;\sim\;
    \mathcal O\!\big(m^d \log m^d + m^d\big)
    \;\sim\;
    \mathcal O\!\big(m^d \log m^d\big).
\end{equation}
The activation memory footprint is dominated by storing $\mat u$, $\ten F_m \mat u$, the pre–activation
$\mat z(\mat u,\cdot)$ and the post–activation:
\begin{equation}\label{eq:fno-activations}
    \text{memory}\big(\ten L_{\text{\tiny FNO}}\big)
    \;=\;
    \mathcal O\!\big(m^d\,C_{\text{in}} + m^d\,C_{\text{out}}\big)
    \;\sim\;
    \mathcal O(m^d).
\end{equation}
The parameter memory is $\mathcal O\!\big(m^d\,C_{\text{in}} C_{\text{out}}\big)$ for the spectral multipliers plus
$\mathcal O\!\big(C_{\text{in}} C_{\text{out}}\big)$ for $\mat W_\theta$.

\paragraph{KANO layer.}
Now consider a KANO layer~(\ref{eq:kano-layer})
\[
  \ten L_{\text{\tiny{KANO}}}(\mat{a})(\mat{x})
  \;=\;
  \bm{\Phi}_{\theta_1}\!\Bigl(
     \ten F_m^{-1}\!\bigl[\,\boldsymbol{p}_{\theta_2}(\mat{x},\bm{\xi})\,*\,\ten F_m (\mat{a})(\bm{\xi})\,\bigr](\mat{x})\;,\;
     \mat{a}(\mat{x})
  \Bigr),
\]
where $\boldsymbol{p}_{\theta_2}(\mat x,\bm{\xi})$ is implemented by a width–fixed KAN symbol network and $\bm{\Phi}_{\theta_1}$ is another width–fixed KAN activation network. Using Kohn–Nirenberg quantization~(\ref{eq:kn_quant}):
\[
      \ten F_m^{-1}\!\bigl[\,\boldsymbol{p}(\mat{x},\bm{\xi})\,*\,\ten F_m (\mat{a})(\bm{\xi})\,\bigr](\mat{x})
  :=
  \Bigl(\frac{h}{L}\Bigr)^d
  \sum_{\bm{\xi}\in\Xi} \sum_{\mat{y}\in\mathcal{Y}}
        e^{i(\mat{x}-\mat{y})\cdot\bm{\xi}}\;
        \boldsymbol{p}(\mat{x},\bm{\xi})\,
        \mat{a}(\mat{y}),
\]
on a $d$-dimensional grid $\mathcal Y$ with $m^d$ points and a retained frequency set $\Xi$ with $m^d$ modes, a single forward application of $\ten L_{\text{\tiny KANO}}$ has the following costs:
\begin{itemize}
    \item \emph{KN quantization (double sum):}
    for each output location $\mat x$ we evaluate the double sum over
    $\bm\xi\in\Xi$ and $\mat y\in\mathcal Y$, i.e.\ $m^{2d}$ terms per $\mat x$.
    Each term involves a dense matrix–vector product
    $\boldsymbol{p}(\mat{x},\bm{\xi})\,\mat a(\mat y)$ of cost
    $\mathcal O(C_{\text{in}} C_{\text{out}})$.
    Summed over all $\mat x\in\mathcal Y$ this yields
    \[
        \text{KN operator} \;\sim\; \mathcal O\!\big(m^{3d}\,C_{\text{in}} C_{\text{out}}\big).
    \]
    \item \emph{Symbol network evaluation:}
    the symbol KAN $\boldsymbol{p}_{\theta_2}$ is evaluated once per pair
    $(\mat x,\bm\xi)\in\mathcal Y\times\Xi$, i.e.\ at $m^{2d}$ points.
    Let $P_p$ denote the cost of a single forward evaluation of $\boldsymbol{p}_{\theta_2}$.
    Then
    \[
        \text{symbol KAN} \;\sim\; \mathcal O\!\big(m^{2d}\,P_p\big).
    \]
    \item \emph{Activation KAN:}
    the activation network $\bm{\Phi}_{\theta_1}$ is applied point-wise at each spatial point $\mat x\in\mathcal Y$.
    Let $P_\Phi$ denote the cost of one forward pass of $\bm{\Phi}_{\theta_1}$.
    Then
    \[
        \text{activation KAN} \;\sim\; \mathcal O\!\big(m^d\,P_\Phi\big).
    \]
\end{itemize}
Collecting terms, for fixed channel counts and fixed KAN architectures we obtain the per–layer forward cost:
\begin{equation}\label{eq:kano-flops}
    \text{FLOPs}\big(\ten L_{\text{\tiny KANO}}\big)
    \;\sim\;
    \mathcal O\!\big(m^{3d}\,C_{\text{in}} C_{\text{out}} + m^{2d} P_p + m P_\Phi\big).
\end{equation}
The activation memory footprint is dominated by storing the feature maps and, if materialized, the symbol grid $\boldsymbol{p}_{\theta_2}(\mat x,\bm\xi)$:
\begin{equation}\label{eq:kano-activations}
    \text{memory}\big(\ten L_{\text{\tiny KANO}}\big)
    \;\sim\;
    \mathcal O\!\big(m^d\,C_{\text{in}} + m^d\,C_{\text{out}} + m^{2d}\,C_{\text{in}} C_{\text{out}}\big)
    \;\sim\;
    \mathcal O\!\big(m^{2d}\,C_{\text{in}} C_{\text{out}}\big),
\end{equation}
in addition to the parameter size of the KAN subnetworks.

\paragraph{Inference cost of compact KAN subnetworks.}
For completeness, we quantify the inference cost $P_p$ and $P_\Phi$ of the KAN subnetworks.
Consider a fully connected KAN layer with input width $d_{\text{in}}$, output width $d_{\text{out}}$,
and $G$ basis functions per edge. Each edge $(j\to i)$ carries a learnable univariate function
$f_{ij}:\mathbb R\to\mathbb R$ represented as
\[
  f_{ij}(x) \;=\; \sum_{g=1}^G w_{ijg}\,\phi_g(x),
\]
where $\{\phi_g\}_{g=1}^G$ are fixed basis functions (e.g.\ B–splines or rational functions)
and $w_{ijg}$ are learned coefficients. Evaluating $f_{ij}(x_j)$ for a given scalar input $x_j$
requires computing the active basis functions $\phi_g(x_j)$ and a dot product over $G$ elements.

A single KAN layer thus computes, for each output coordinate $i$,
\[
  y_i \;=\; \sum_{j=1}^{d_{\text{in}}} f_{ij}(x_j),
\]
and the total cost of one forward pass through this layer would be $\mathcal O\!\big(d_{\text{out}}\,d_{\text{in}}\,G\big)$, up to lower order terms from basis evaluation. The parameter count of this layer is of the same order, $\mathcal O(d_{\text{out}}\,d_{\text{in}}\,G)$.
Under the assumption of compact KAN subnetworks, widths, depths, and the number of basis functions $G$ are all small and independent to the operator resolution $m$. If we denote by $N$ an upper bound on their layer widths and by $L_{\text{\tiny KAN}}$ their depth,
then their total inference costs satisfy
\[
  P_p,\;P_\Phi
  \;=\;
  \mathcal O\!\big(L_{\text{\tiny KAN}}\,N^2\,G\big).
\]
In the per–layer KANO complexity~(\ref{eq:kano-flops}) the KAN subnetworks contribute only a constant–factor overhead that does not grow with the spatial or spectral resolution, hence for large resolution $m\gg N, L_{\text{\tiny KAN}}$ the dominant term would be the KN quantization term $\mathcal O(m^{3d} C_{\text{in}} C_{\text{out}})$.

\subsection{Computation Complexity as a Function of Accuracy for Position-Dependent Dynamics}

As apparent in Eq.~(\ref{eq:fno-flops} and Eq.~(\ref{eq:kano-flops}), KANO requires orders heavier FLOPs compared to FNO due to the expensive double sum nature of its KN quantization, when two models are of similar size. For small resolution $m$, the inference cost $P_p$ and $P_\Phi$ still set hard lower bound on the computation complexity as well. However, one of the main arguments we make in this work is that KANO enjoys incomparable parameter efficiency compared to FNO on the target class of spectrally dense operators such as position-dependent dynamics: parameter complexity of KANO scales polynomially where that of FNO can scale super-exponentially on a spectrally dense ground-truth operator. Therefore, KANO's expensive FLOPs requirement can be compensated by parameter efficiency where FNO suffers the curse of dimensionality.

For a position-dependent dynamics $\ten G$ and accuracy $\varepsilon>0$, we write
$|\Theta_{\text{\tiny FNO}}(\varepsilon)|$ and $|\Theta_{\text{\tiny KANO}}(\varepsilon)|$ for the
smallest parameter counts required by FNO and KANO respectively to achieve
$\|\ten G - \ten G_\theta\| \leq \varepsilon$, following Theorem~\ref{thm:fno-fail} and
Theorem~\ref{thm:kano-general}. Also, let $T_{\text{\tiny FNO}}(\varepsilon)$ and $T_{\text{\tiny KANO}}(\varepsilon)$ denote the cost of one
forward (or forward+backward) pass of an FNO or KANO architecture chosen to achieve accuracy
$\varepsilon$ on the ground-truth operator. We do not attempt to model the number of optimization steps;
instead, we focus on how the cost of a \emph{single} training or inference step scales with $\varepsilon$. Assuming a model must perform at least a constant number of FLOPs per parameter in each step, lower bound for computation complexity can be estimated as:
\begin{equation}\label{eq:comp_accuracy}
    T_{\text{\tiny FNO}}(\varepsilon)
  \;\gtrsim\;
  c_0\,|\Theta_{\text{\tiny FNO}}(\varepsilon)|, \qquad  T_{\text{\tiny KANO}}(\varepsilon)
  \;\gtrsim\;
  c'_0\,|\Theta_{\text{\tiny KANO}}(\varepsilon)|
\end{equation}
for constants $c_0, c'_0>0$. Such lower bound assumption is not universal for any neural network, but in FNO architecture all its parameters equally appear in the matrix computation via the dense parameter block $\mat R$ and $\mat W$; especially for implementations that evaluate and update all parameters in every step. Hence, computation complexity lower bound assumption by its model size is reasonable for FNO. 

\paragraph{Computation complexity of FNO by accuracy for position operators from Theorem~\ref{thm:fno-fail}.}
For the position–multiplier
$\ten M(\mat x) := x_1^{\alpha_1}\cdots x_d^{\alpha_d}$ of total degree $M := |\alpha|\ge 1$, the
restatement of Theorem~\ref{thm:fno-fail} in Appendix~\ref{app:thm1proof} shows that an FNO
$\ten G^{\text{\tiny FNO}}_\theta$ that achieves
\[
  \|\ten M - \ten G^{\text{\tiny FNO}}_\theta\|_{H^s} \le \varepsilon
\]
on band–limited inputs can suffer curse of dimensionality:
\begin{equation}\label{eq:fno-param-scaling}
  |\Theta_{\text{\tiny FNO}}(\varepsilon)|
  \;\sim\;
  \exp\!\bigl(c\,\varepsilon^{-M d/s}\bigr),
\end{equation}
for some constant $c>0$ depending only on geometric constants of Sobolev space $H^s$. From Eq.~(\ref{eq:comp_accuracy}), the computation complexity of FNO by accuracy on $\ten M$ can be written as:
\begin{equation}\label{eq:fno_accuracy_comp}
    T_{\text{\tiny FNO}}(\varepsilon)
  \;\gtrsim\;
  c_0\,|\Theta_{\text{\tiny FNO}}(\varepsilon)|\sim \mathcal{O}\bigl(\exp\!(\varepsilon^{-M d/s})\bigr)
\end{equation}

\paragraph{Computation complexity of KANO by accuracy for position operators from Theorem~\ref{thm:kano-general}.}
On the other hand, Theorem~\ref{thm:kano-general} and Corollary~\ref{cor:finite-pm} show that as the KANO projection of $\ten M$ yields symbols $\mat p_{\ten M}(\mat x,\bm\xi)$ of sufficient smoothness, KANO can
achieve $\|\ten M - \ten G^{\text{\tiny KANO}}_\theta\|\le\varepsilon$ with
\begin{equation}\label{eq:kano-param-scaling}
  |\Theta_{\text{\tiny KANO}}(\varepsilon)|
  \;\sim\;
  \mathcal O\!\bigl(\varepsilon^{-\beta}\bigr),
\end{equation}
where $\beta = d/(2s_p)$ or $\beta = d/(2r)$ is a geometric exponent determined by the symbol regularity~\citep{expressivitykan}. From Eq.~(\ref{eq:comp_accuracy}), the computation complexity of KANO by accuracy on $\ten M$ can be written as:
\begin{equation}\label{eq:kano_accuracy_comp}
    T_{\text{\tiny KANO}}(\varepsilon)
  \;\gtrsim\;
  c'_0\,|\Theta_{\text{\tiny KANO}}(\varepsilon)|\sim \mathcal{O}\bigl(\exp\!(\varepsilon^{-\beta})\bigr)
\end{equation}

Therefore, when the ground-truth operator is spectrally dense yet its KANO projection give smooth enough symbol so that both Theorem~\ref{thm:fno-fail} and Theorem~\ref{thm:kano-general} are effective (for instance, position-dependent dynamics), lower bound of computation complexity of FNO can prevail over that of KANO because of the curse of dimensionality discussed in Section~\ref{sec:thm1}.

We emphasize that this is an asymptotic statement under the specific operator class where both Theorem~\ref{thm:fno-fail} and Theorem~\ref{thm:kano-general} hold. For operators that are nearly spectrally diagonal (e.g. standard FNO benchmarks), FNO is theoretically much faster than KANO when the two models are in similar size. However, when learning a spectrally dense, variable–coefficient operators to high precision, KANO's parameter efficiency can, in theory, compensate for its expensive double sum KN quantization while FNO
faces a super–exponential growth in model size hence per–step FLOPs to achieve the same level of accuracy.

%% file: appendix_llm.tex
Large Language Model (LLM) is used to polish the writing in this paper, such as detecting grammar errors and typos. LLM is also used to search for appropriate references for correct citations, and all the proposed references are fully inspected and verified before citing in the paper.